\documentclass[runningheads]{llncs}
\usepackage{graphicx}

\usepackage{tikz}
\usepackage{comment} 
\usepackage{amsmath,amssymb} 
\usepackage{color}

\usepackage[width=122mm,left=12mm,paperwidth=146mm,height=193mm,top=12mm,paperheight=217mm]{geometry}

\usepackage{url}
\usepackage{graphicx,wrapfig}
\usepackage{booktabs,multirow}
\usepackage{colortbl}
\definecolor{hui}{gray}{0.3}
\definecolor{ggray}{gray}{0.85}
\newcolumntype{a}{>{\columncolor{ggray}}c}
\usepackage[ruled]{algorithm2e}
\usepackage[pagebackref=true,colorlinks,bookmarks=false]{hyperref}

\begin{document}
	\pagestyle{headings}
	\mainmatter
	\def\ECCVSubNumber{1294}  
	\title{A Balanced and Uncertainty-aware Approach for Partial Domain Adaptation\thanks{J. Feng was partially supported by NUS ECRA FY17 P08, AISG-100E2019-035, and MOE Tier 2 MOE2017-T2-2-151. The authors also thank Quanhong Fu for her help to improve the technical writing aspect of this paper.}}

	\titlerunning{A Balanced and Uncertainty-aware Approach for Partial Domain Adaptation}
	%
	\author{Jian Liang\inst{1}\orcidID{0000-0003-3890-1894} \and Yunbo Wang\inst{2}\orcidID{0000-0002-6215-8888} \and Dapeng Hu\inst{1} \and Ran He\inst{3}\orcidID{0000-0002-3807-991X} \and Jiashi Feng\inst{1}\orcidID{0000-0001-6843-0064}}
	\authorrunning{Liang et al.}
	%
	\institute{Department of ECE, National University of Singapore (NUS)\\
	\email{liangjian92@gmail.com},
	\quad\email{dapeng.hu@u.nus.edu},
	\quad\email{elefjia@nus.edu.sg}\\
	\and
	Peking University
	\quad\email{wangyunbo09@gmail.com}
	\and
	Institute of Automation, Chinese Academy of Sciences
	\quad\email{rhe@nlpr.ia.ac.cn}
	}
	\maketitle

\begin{abstract}
This work addresses the unsupervised domain adaptation problem, especially in the case of class labels in the target domain being only a subset of those in the source domain. 
Such a partial transfer setting is realistic but challenging and existing methods always suffer from two key problems, negative transfer and uncertainty propagation. 
In this paper, we build on domain adversarial learning and propose a novel domain adaptation method BA$^3$US with two new techniques termed Balanced Adversarial Alignment (BAA) and Adaptive Uncertainty Suppression (AUS), respectively.
On one hand, negative transfer results in misclassification of target samples to the classes only present in the source domain.
To address this issue, BAA pursues the balance between label distributions across domains in a fairly simple manner. 
Specifically, it randomly leverages a few source samples to augment the smaller target domain during domain alignment so that classes in different domains are symmetric. 
On the other hand, a source sample would be denoted as uncertain if there is an incorrect class that has a relatively high prediction score, and
such uncertainty easily propagates to unlabeled target data around it during alignment, which severely deteriorates adaptation performance. 
Thus we present AUS that emphasizes uncertain samples and exploits an adaptive weighted complement entropy objective to encourage incorrect classes to have uniform and low prediction scores.
Experimental results on multiple benchmarks demonstrate our BA$^3$US surpasses state-of-the-arts for partial domain adaptation tasks. 
Code is available at \url{https://github.com/tim-learn/BA3US}.
\keywords{Partial Transfer Learning; Domain Adaptation; Adversarial Alignment; Uncertainty Propagation; Object Recognition.}
\end{abstract}

\section{Introduction}
Over the past two decades, many research efforts have been devoted to unsupervised domain adaptation (UDA), which aims to leverage labeled source domain data to learn to classify unlabeled target domain data.
Typically, existing UDA methods minimize the discrepancy between two domains by matching their statistical distribution moments \cite{tzeng2014deep,sun2016deep,zellinger2016central,koniusz2017domain} or by domain adversarial learning~\cite{tzeng2015simultaneous,ganin2016domain,tzeng2017adversarial,long2018conditional}. 
Once the domain shift is mitigated, source classifiers can be easily transferred to the target domain even with no labeled target domain data available. 
However, both UDA strategies always assume that different domains share the same label space. 
Such an assumption may not hold in practice and target domain labels may be only a subset of source domain labels.
This introduces an unsupervised partial domain adaptation (PDA) problem that receives increasing research attention recently \cite{cao2018partialb,zhang2018importance,cao2018partial,matsuura2018twins}.

The PDA problem is challenging since source-only classes may occur in the target domain during distribution alignment, which is well-known as class mismatch that potentially causes \textit{negative transfer}. 
Several previous PDA approaches \cite{cao2018partialb,cao2018partial} mitigate negative transfer by jointly filtering out source-only classes and promote positive transfer by matching the data distribution in the shared classes.
Samples from source-only classes are expected to have lower weights in the adaptation module such that the marginal distributions of two domains can be aligned well.
However, it is rather risky to rule out the source-only classes, especially when the estimation of label distribution in the target domain is inaccurate.

To match two non-identical label spaces, we view the PDA problem from a new perspective and propose to augment the target label space to be the same as the source label space.
Specifically, we develop a simple balanced alignment solution, termed Balanced Adversarial Alignment (BAA), that borrows fewer and fewer samples from the source domain to the target domain within an iterative adversarial learning framework.
We expect that the augmented target domain looks much more similar to the source domain w.r.t. the label distribution, and the challenging PDA problem can be transformed to a well-studied UDA task.
To focus on the originally shared classes, we propose to filter out the source-only classes via a class-level weighting scheme meanwhile, making the large UDA task more compact.

Besides, existing domain adaptation methods always employ a conventional cross-entropy loss to merely promote the prediction score of ground-truth classes but neglect to suppress those of incorrect classes, which may result in a new problem termed \textit{uncertainty propagation}.
Intuitively, if incorrect classes have relatively high prediction scores for source data, some wrong classes would possibly have the largest prediction scores for the aligned target data around them.
This problem is quite critical but has been always ignored in the domain adaptation field.	
To circumvent the issue, we develop an uncertainty suppression solution termed Adaptive Uncertainty Suppression (AUS) that exploits complement entropy \cite{chen2019complement} in the labeled source domain to prohibit possibly high prediction scores from incorrect classes.
Specifically, we emphasize more the uncertain samples corresponding to smaller cross-entropy loss (confidence) and propose a confidence-weighted complement entropy objective in addition to the primary cross-entropy objective.

Generally, our baseline is built on the seminal domain adversarial networks \cite{ganin2015unsupervised,ganin2016domain} and exploits conditional entropy minimization and an entropy-aware weight strategy~\cite{long2018conditional}.
In this paper, we equip the baseline model with two proposed techniques mentioned above and finally formulate a unified framework BA$^3$US which well addresses the negative transfer and uncertainty propagation problems in partial domain adaptation.
We also empirically discover that the uncertainty suppression technique works well for vanilla closed-set domain adaptation.

To sum up, we make the following contributions. To our best knowledge, this is the first work that tackles partial domain adaptation by augmenting the target domain and transforming it into a UDA-like problem. 
The proposed balanced augmentation technique is fairly simple and works very well for PDA tasks.
Besides, we address an overlooked issue in this field named uncertainty propagation by designing an adaptive weighted complement entropy for the source domain.
Extensive results demonstrate that our approach yields new state-of-the-art results on several visual benchmark datasets, including Office31~\cite{saenko2010adapting}, Office-Home~\cite{venkateswara2017deep}, and ImageNet-Caltech~\cite{cao2019learning}.

\section{Related Work}
The past two decades have witnessed remarkable progress in domain adaptation. Interested readers can refer to \cite{csurka2017domain,zhang2019transfer,kouw2019review} for taxonomy and survey.

\textbf{Unsupervised Domain Adaptation (UDA).} 
Compared with its supervised counterpart, UDA is more practical and challenging since no labeled data in the target domain are available. 
Recently, deep convolutional neural networks have achieved great success for visual recognition tasks, and we focus on deep UDA methods in this work. They can be categorized into three main groups.
The first group aims to minimize the domain discrepancy by matching different statistic moments like maximum mean discrepancy (MMD) \cite{tzeng2014deep,long2015learning,long2017deep,liang2019aggregating} and higher-order moment matching \cite{sun2016deep,zellinger2016central,koniusz2017domain}.
The second group that is widely used introduces a domain discriminator and exploits the idea of adversarial learning \cite{goodfellow2014generative} to encourage domain confusion so that the discriminator can not decide which domain the data come from. 
Some typical examples are \cite{tzeng2015simultaneous,ganin2016domain,tzeng2017adversarial,bousmalis2017unsupervised}.
A third group is a reconstruction-based approach, assuming the reconstruction of both source and target domain samples to be important and helpful. Among them,~\cite{ghifary2016deep,zhu2017unpaired} utilize encoder-decoder reconstruction and adversarial reconstruction, respectively. 
Despite their success for vanilla UDA, they are easily stuck by negative transfer for PDA due to the mismatched marginal label distributions.

\textbf{Partial Domain Adaptation (PDA).}
In reality, PDA can be considered as a special case of imbalanced domain adaptation, where the target label distribution is quite dissimilar to that of the source domain.
Until recent years,~\cite{ming2015unsupervised} first introduces an imbalanced scenario where label numbers of the source and target domains are not the same, which draws the attention of many researchers~\cite{tsai2016domain,cao2018partialb,cao2018partial,zhang2018importance,matsuura2018twins,cao2019learning,hu2019multi}.
Different from shallow methods \cite{ming2015unsupervised,tsai2016domain}, deep methods \cite{cao2018partialb,cao2018partial,zhang2018importance,matsuura2018twins} are mainly based on the domain adversarial learning framework and achieve promising recognition accuracy.
Selective adversarial network (SAN)~\cite{cao2018partialb} exploits a multi-discriminator domain adversarial network and tries to select source-only classes by imposing different localized weights on different discriminators.
Importance weighted adversarial nets (IWAN)~\cite{zhang2018importance} apply only one domain discriminator and weigh each source sample with the probability of being a target sample.
Partial adversarial domain adaptation (PADA)~\cite{cao2018partial} estimates the target label distribution and then feeds the class-wise weights to both the source discriminator and the domain discriminator, while two weighted inconsistency-reduced networks (TWINs)~\cite{matsuura2018twins} leverage two independent networks to estimate the target label distribution and minimize the domain difference measured by the classifiers' inconsistency on the target samples. 
Deep Residual Correction Network (DRCN) \cite{li2020deep} proposes a weighted class-wise matching strategy to explicitly align target data with the most relevant source subclasses.
Recently, Example Transfer Network (ETN)~\cite{cao2019learning} jointly learns domain-invariant representations across domains and a progressive weighting scheme to quantify the transferability of source examples, which achieves state-of-the-art results on several benchmark datasets.
Generally, all the PDA methods above attempt to filter out the large source domain to match the small target domain. Comparatively, our method tries to augment the small target domain to match the source domain from a different perspective.

\textbf{Data Synthesis and Augmentation.}
Recently, synthesis and augmentation techniques like CycleGAN \cite{zhu2017unpaired} and \emph{mixup} \cite{zhang2018mixup} are favored by UDA and semi-supervised learning methods for improving performance.
For example, \cite{hoffman2018cycada} directly exploits CycleGAN to generate target-like images from source samples to narrow the domain shift for adaptive semantic segmentation.
\cite{mao2019virtual,wang2019semi} extend \emph{mixup} to domain adaptation and generate pseudo training samples via interpolating between certain source samples and uncertain target samples.
To some degree, target augmentation in this paper is like a special case of \emph{mixup} where the mixup coefficient is always binary. 
However, the motivations are totally different. Our method considers neither the interpolated semantic label nor the interpolated domain label for a classification loss.

\section{Proposed Method}
We elaborate on the proposed framework for partial domain adaptation (PDA) in this section.
First, we give definitions and notations. 
We follow the protocol of unsupervised PDA where we have a labeled source domain dataset $\mathcal{D}_s = \{(x_i^s,y_i^s)\}_{i=1}^{n_s}, x_i^s \in \mathbb{R}^{d}$ and an unlabeled target domain dataset $\mathcal{D}_t = \{(x_i^t)\}_{i=1}^{n_t}, x_i^t \in \mathbb{R}^{d}$ during the training stage.
These two domains have different feature distributions: $p_s(x_s) \not = p_t(x_t)$ due to the domain shift.
Notably, different from vanilla UDA, the target labels are a subset of the source labels for PDA: $\mathcal{Y}_t \subseteq \mathcal{Y}_s$, and $C$ denotes the total number of classes in $\mathcal{Y}_s$.

We aim to learn a deep neural network $h: \mathcal{X}\to \mathcal{Y}$ that consists of two components: $h = g \circ f$. Here $f: \mathcal{X}\to \mathcal{Z}$ denotes the feature extractor and $g: \mathcal{Z}\to \mathcal{Y}$ denotes a class predictor.
Since we target at learning domain-invariant features, the prediction function is assumed identical, i.e., $g=g_s=g_t$.
For simplicity, we also share the feature extractor $f$ for different domains.
We introduce an adversarial classifier $D: \mathcal{Z}\to \{0,1\}$ to mitigate the distribution discrepancy across domains as explained later.

\subsection{Domain Adversarial Learning Revisited}
Generative adversarial network (GAN) \cite{goodfellow2014generative} learns two competing components: the discriminator $D$ and the generator $F$ which play a minimax two-player game, where $F$ tries to fool $D$ by generating examples that are as realistic as possible and $D$ tries to classify the generated samples as fake.
Inspired by the idea of GAN, domain adversarial neural networks (DANN) \cite{ganin2015unsupervised,ganin2016domain} develops a two-player game for UDA, where one player $D(\cdot;\theta_d)$ (i.e., domain discriminator) tries to distinguish the source domain datum from that of the target domain and the other player $F(\cdot;\theta_f)$ (i.e., feature extractor) is trained to confuse the domain discriminator $D(\cdot;\theta_d)$.
Generally, the minimax game of DANN is formulated as
\begin{equation}
\begin{aligned}
\min\limits_{\theta_f, \theta_g} \max\limits_{\theta_d} &\ \mathcal{L}_{cls}(\theta_f, \theta_g) + \lambda \ \mathcal{L}_{adv}(\theta_f, \theta_d),\\
\mathcal{L}_{adv}(\theta_f, \theta_d) &= \frac{1}{n_s}\sum\nolimits_{i=1}^{n_s} \log[D(F(x_i^s))]+ \frac{1}{n_t} \sum\nolimits_{j=1}^{n_t} \log[1 - D(F(x_j^t))], \\
\mathcal{L}_{cls}(\theta_f, \theta_g) &= \frac{1}{n_s} \sum\nolimits_{i=1}^{n_s} {l}_{ce}(G(F(x_i^s)), y_i^s),
\end{aligned}
\label{dann}
\end{equation}
where ${l}_{ce}(\cdot, \cdot)$ represents the softmax cross-entropy loss, $G(\cdot; \theta_g)$ denotes the source classifier, $F(\cdot;\theta_f)$ represents the domain-shared feature extractor, and $\lambda$ is a hyper-parameter to trade-off the source risk and domain adversary.
Different from a label flipping step in GAN, a gradient reversal layer (GRL) is further defined in \cite{ganin2015unsupervised} to optimize the objective in Eq.~(\ref{dann}). 
Due to its simplicity and effectiveness, the idea of domain adversarial learning has been adopted in many previous domain adaptation works \cite{liu2016coupled,tzeng2017adversarial,long2018conditional}.

As shown in Eq.~(\ref{dann}), each sample from both source and target domains is equally involved in the adversarial loss $\mathcal{L}_{adv}$, which seems not reasonable.
If we only have samples distributed in the margin of the classifier (called `hard' samples) from different domains and pursue domain alignment via them, it may perform badly for those `easy' samples across domains.
As such, we expect those `hard' samples and `easy' samples to own lower and higher weights during domain adversarial alignment.
Specifically, we quantify the difficulty via the entropy criterion $H(h) = -\sum_{c=1}^{C}h_c\log(h_c)$ and adopt the same weighting strategy as \cite{long2018conditional} to impose an entropy-aware weight $w(x) = 1 + e^{-H(h(x))}$ on each sample,
\begin{equation}
\begin{aligned}
\mathcal{L}_{adv}^{e}(\theta_f, \theta_d) = \frac{1}{n_s}\sum\nolimits_{i=1}^{n_s} w(x_i^s)\log[D(F(x_i^s))] + \frac{1}{n_t}\sum\nolimits_{j=1}^{n_t} w(x_j^t)\log[1 - D(F(x_j^t))].
\end{aligned}
\end{equation}

Obviously, if there is no domain shift, UDA (including PDA) degenerates to a typical semi-supervised learning problem.
On one hand, we aim to mitigate the domain shift; on the other hand, this motivates us to adopt the popular entropy minimization principle \cite{grandvalet2005semi} in semi-supervised learning to the PDA task.
It is desirable that all the unlabeled target samples have highly-confident predictions. This is encouraged via the following conditional entropy term,
\begin{equation}
\mathcal{L}_{ent}(\theta_f,\theta_y) =     \frac{1}{n_t}\sum\nolimits_{j=1}^{n_t} H(G(F(x_j^t))).
\label{tar_ent}
\end{equation}
Inspired by the observation in \cite{cao2018partialb} that redundant information is not beneficial for adaptation, we adopt the following optimization objective of Entropy-regularized DANN (\textbf{E-DANN}) as an initial model of our method,
\begin{equation}
\min\limits_{\theta_f, \theta_g} \max\limits_{\theta_d} \ \mathcal{L}_{cls}^{w}(\theta_f,\theta_g) + \alpha \mathcal{L}_{ent}(\theta_f,\theta_g) + \lambda \mathcal{L}_{adv}^{e}(\theta_f,\theta_d),
\label{edann}
\end{equation}
where $\mathcal{L}_{cls}^{w}(\theta_f, \theta_g)=\frac{1}{n_s} \sum\nolimits_{i=1}^{n_s} m(y_i^s) {l}_{ce}(G(F(x_i^s)), y_i^s)$, $m$ denotes the normalized estimated class-level weight vector via the target domain, and $\alpha,\lambda$ are two empirical trade-off parameters.
	
\begin{figure*}[t]
	\centering
	\includegraphics[width=0.9\linewidth, trim=10 20 0 35,clip]{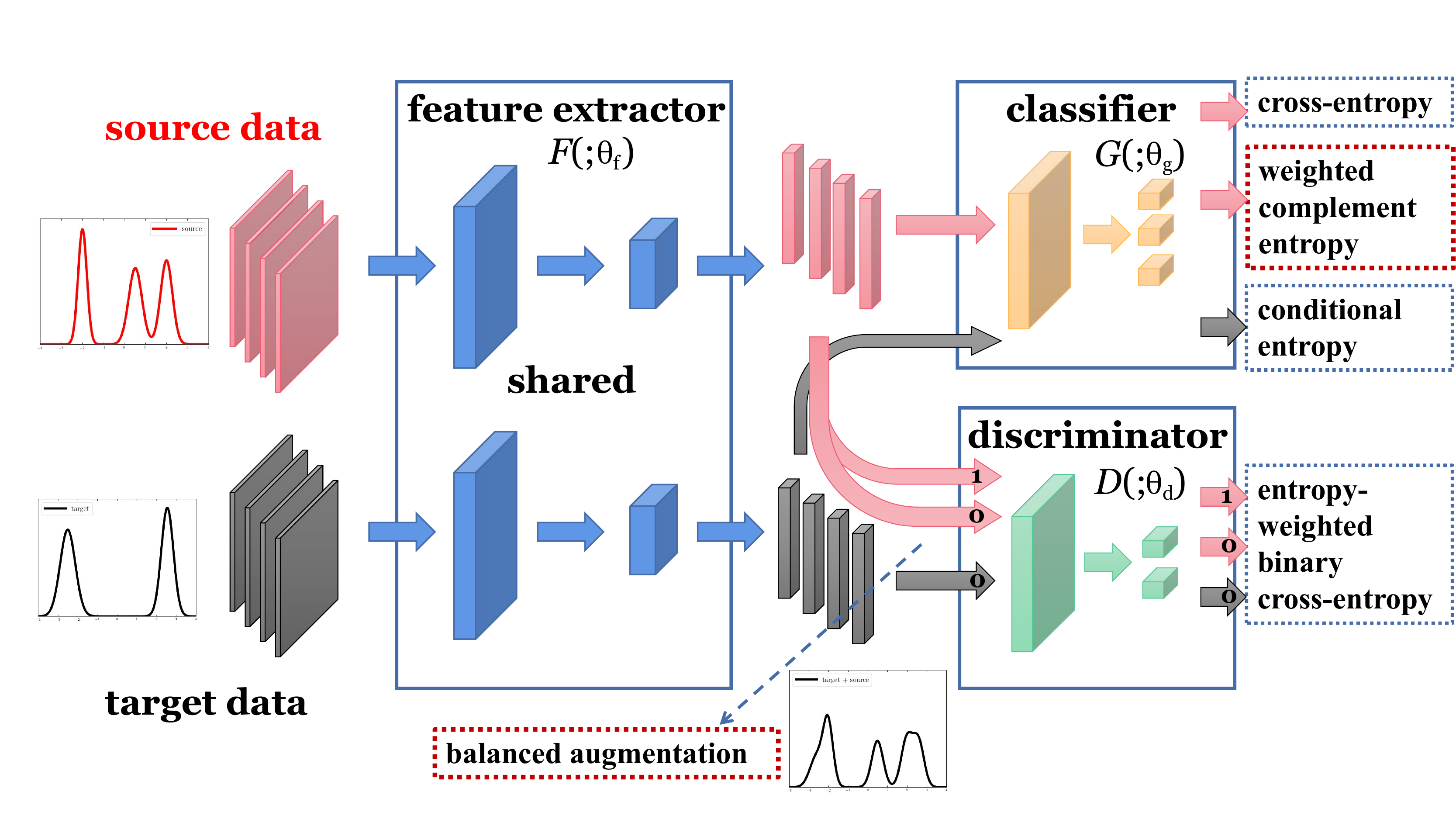}
	\caption{Architecture of our domain adaptation method. There are three modules: a shared feature extractor $F$, a classifier $G$ and a domain discriminator $D$. Different from domain adversarial learning \cite{ganin2015unsupervised}, it contains two extra components with marked red border, i.e., balanced augmentation and weighted complement entropy.}
	\label{fig:framework}
\end{figure*}
	
\subsection{Balanced Adversarial Alignment (BAA)}
\label{sec:baa}
The joint distribution shift is the actual root to negative transfer \cite{wang2019characterizing}. 
For example, the marginal label distributions are not symmetric in PDA, and thus source-only classes are prone to be matched with target classes, resulting in a negative transfer problem. 
Generally, a class-level source re-weighting scheme sounds a natural choice for PDA, since it is expected to filter out source-only classes and promote positive transfer between the shared classes across domains.
Previous methods \cite{cao2018partial,matsuura2018twins} resort to target predictions to generate class-level weights and effectively avoid negative transfer to some degree.
Ideally, PDA with a weighting scheme behaves like a small UDA problem, but it heavily relies on accurate target predictions to calculate suitable weights.

In contrast, we propose an extremely simple scheme as shown in Fig.~\ref{fig:framework} for the challenging distribution alignment in PDA.
Intuitively, we pursue the balance between different label distributions across domains with quite an opposite solution, i.e., augmenting the target domain using original source samples instead of weighting the source domain.
This idea looks weird but is actually reasonable because we readily turn the PDA task into a large UDA-like task and the negative transfer effects caused by source-only classes can be well alleviated. 
The detailed formulation of balanced augmentation (alignment) is shown below,
\begin{equation}
\begin{aligned}
& \qquad\qquad\qquad \mathcal{L}_{adv}^{ba}(\theta_f,\theta_d) = \frac{1}{n_s} \sum\nolimits_{i=1}^{n_s} w(x_i^s) m(y_i^s) \log[D(F(x_i^s))]\ +\\ 
& \frac{1}{n_t} \sum\nolimits_{j=1}^{n_t} w(x_j^t)\log[1 - D(F(x_j^t))] + \frac{\rho}{n_s} \sum\nolimits_{i=1}^{n_s} w(x_i^s) m(y_i^s) \log[1-D(F(x_i^s))].
\end{aligned}
\label{aug}
\end{equation}
Specifically, we adopt a progressive strategy for the target augmentation scheme. We borrow the source samples with different ratios, and the ratio $\rho$ gradually decreases to 0 as the number of iterations increases.
This is because the learned feature representations in early iterations are not quite transferable and we need more source samples to avoid class mismatch.
When the learned features are desirably discriminative and transferable, the estimation of class-level weights becomes more accurate, making the augmentation trivial.

Note that we borrow original samples from the source domain rather than exploiting a generative model like CycleGAN~\cite{zhu2017unpaired} to synthesize target-like source images. 
The reason is that obtaining data-dependent translation models between such heterogeneous domains is quite time-consuming and we find translation does not even improve the adaptation results.

\subsection{Adaptive Uncertainty Suppression (AUS)}
	\label{sec:aus}
	\begin{wrapfigure}{r}{0.5\textwidth}
		\centering
		\includegraphics[width=1.0\linewidth, trim=180 280 95 80,clip]{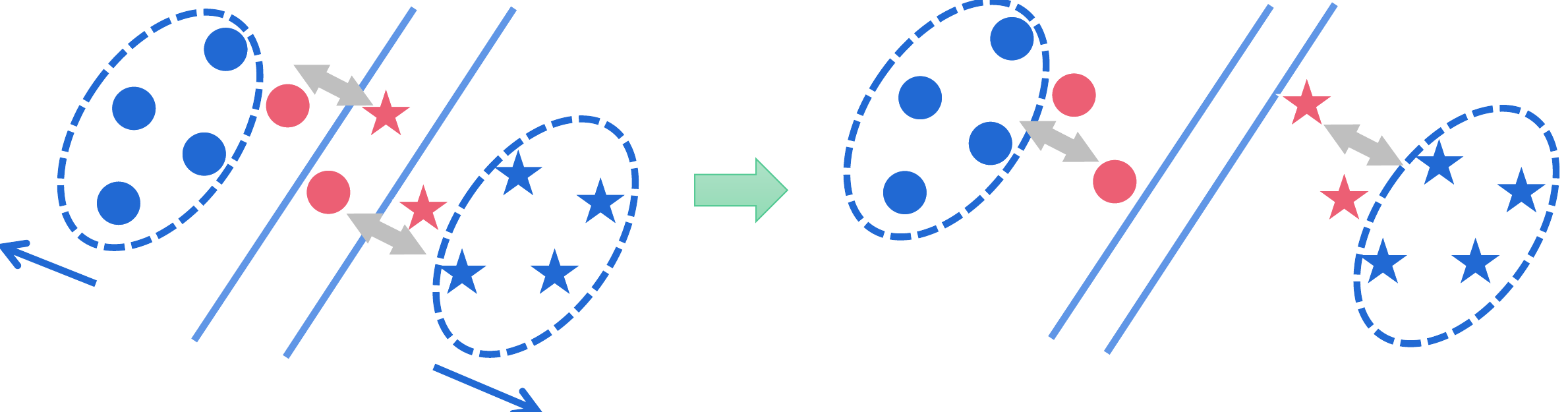}
		\caption{Mitigating the effects of uncertainty propagation from source. [\textcolor{blue}{blue}: source, \textcolor{red}{red}: target, \textcolor{hui}{gray}: adversarial alignment.]}
		\label{fig:wce}
	\end{wrapfigure}
Previous DA methods focus on strengthening the feature transferability by developing various domain alignment strategies, but they mostly ignore the feature discriminability and simply use the conventional cross-entropy loss in the labeled source domain to learn the features.
In that case, even though the domain shift is mitigated, the classifier may perform worse on target data.
This is because source classes are not equally separated from each other, which may lead to the propagation of the confusion (uncertainty) to target predictions and thus confusing features in the target domain, which is termed as the uncertainty propagation problem.
Take a 3-way classification problem as an example. 
There is no class overlap in the source domain, but class 1 and 2 may be close to each other.
A toy example of how class 1 and class 2 behave is shown in Fig.~\ref{fig:wce} where a few samples lie close to the decision boundary between class 1 and class 2.
During domain alignment, some unlabeled target data are enforced to match these source data, which would be easily misclassified. 
However, such a critical problem has always been overlooked in prior literature. 
Looking back at the cross-entropy loss $l_{ce}(\hat{y},y) = -\sum_i y_i \log(\hat{y}_i)$, it only exploits the information from the ground-truth class while ignoring that from other incorrect classes.
For example, the source output \textbf{[0.6, 0.3, 0.1]} is more uncertain than \textbf{[0.6, 0.2, 0.2]}, but both have the same cross-entropy loss.

Though several previous methods \cite{kumar2018co,shu2018dirt} incorporate virtual adversarial training \cite{miyato2018virtual} in the classification term to implicitly increase the margin between different classes, more computation complexity is required and several parameters need to be tuned.
Inspired by \cite{chen2019complement}, we exploit a complement entropy that expects uniform and low prediction scores for incorrect classes for labeled source samples.
To accurately suppress uncertainty, we further place more emphasis on the uncertain samples that own smaller cross-entropy loss (confidence) and propose a confidence-weighted complement entropy objective below,
\begin{equation}
\begin{aligned}
\mathcal{L}_{wce}^{w}(\theta_f, \theta_g) &= \frac{1}{n_s \text{log}(K-1)}\sum\nolimits_{i=1}^{n_s} m(y_i^s) l_{wce}(G(F(x_i^s)), y_i^s),\\
\text{where}\ l_{wce}(\hat{y}, y) &= (1-\hat{y}_a)^\xi \sum\nolimits_{j \not = a} \frac{\hat{y}_j}{1-\hat{y}_a} \log(\frac{\hat{y}_j}{1-\hat{y}_a}),
\end{aligned}
\label{eq:src_coe}
\end{equation}
where $\xi$ is a hyper-parameter and $a$ is the index of ground-truth class~in~$y$.
Different from the complementary training strategy in \cite{chen2019complement}, we exploit the adaptive weighted complement entropy $\mathcal{L}_{wce}$ as a regularizer, which is more efficient.
We also assign a class-level weight for each sample in Eq.~(\ref{eq:src_coe}) like that in $\mathcal{L}_{cls}^{w}$.

\subsection{Unified Minimax Optimization Problem of BA$^3$US}
Finally, we integrate all the terms in Eqs.~(\ref{edann}, \ref{aug}, \ref{eq:src_coe}) on both source and target samples to avoid negative transfer and uncertainty prorogation and derive a unified framework for PDA. The overall min-max objective is formulated as
\begin{equation}
\begin{aligned}
\min\limits_{\theta_f, \theta_g} \max\limits_{\theta_d} \mathcal{L}^w_{cls}(\theta_f,\theta_g) + \alpha \mathcal{L}_{ent}(\theta_f,\theta_g) + \beta \mathcal{L}^w_{wce}(\theta_f, \theta_g) + \lambda \mathcal{L}_{adv}^{ba}(\theta_f,\theta_d),
\end{aligned}
\label{ours}
\end{equation}
where $\beta$ is another trade-off hyper-parameter to balance the complement entropy and the cross-entropy term.
Again, $m \propto \sum\nolimits_{j=1}^{n_t} G(F(x_i^t))$ is the normalized estimated class-level weight vector.
To optimize the objective above, we follow~\cite{ganin2015unsupervised} to introduce a gradient reversal layer and adopt the same progressive strategy for parameter $\lambda$, i.e., increasing $\lambda$ from 0 to 1 as the number of iterations grows.

Our method is closely related to DANN~\cite{ganin2015unsupervised}, sharing similar formalism of the domain adaptation theory \cite{ben2010theory} that the expected target risk $\epsilon_{T}(H)$ on the target examples is bounded by the source risk $\epsilon_{S}(H)$ and other two terms below, 
	\begin{equation}
	\epsilon_{T}(H) \leq \epsilon_{S}(H) + |\epsilon_{S}(H,H^*) - \epsilon_{T}(H,H^*)| + [\epsilon_{S}(H^*) + \epsilon_{T}(H^*)], 
	\label{eq:theory}
	\end{equation}
where $H^* = \arg\min\nolimits_{x\in \mathcal{H}} [\epsilon_{S}(x) + \epsilon_{T}(x)]$ is the ideal joint hypothesis for the combined risk.
DANN further discovers the second term $\mathcal{H}\Delta\mathcal{H}$-distance can be upper bounded by error of the domain adversarial classifier.
As we do not have labels of the target domain, we expect the entropy minimization term on the target domain to help reduce the last term.
Besides, the proposed complement entropy objective alleviates uncertainty propagation to make the weight estimation more accurate, and thus our method would turn PDA into a small UDA task.
In this way, we can expect that our method minimizes the empirical target risk $\textbf{E}_{(x_t,y_t)\sim p_t}[f(x_t)\not = y_t]$.
The detailed optimization is summarized in Algorithm \ref{alg}.

\section{Experiments}
\subsection{Setup}
\textbf{Datasets.} \textbf{Office31} dataset \cite{saenko2010adapting} includes images of 31 object classes from three different domains, i.e., Amazon, DSLR, and Webcam. We follow the standard protocol used in \cite{cao2018partial} and pick up images of 10 categories shared by Office31 and Caltech256~\cite{griffin2007caltech} as target domains. \textbf{Office-Home} dataset  \cite{venkateswara2017deep}  consists of 4 different domains with each containing 65 kinds of everyday objects, i.e., Artistic, Clipart, Product images, and Real World images. Likewise, we follow \cite{cao2018partial} to select the first 25 categories (in alphabetic order) in each domain as a partial target domain. 
\textbf{ImageNet-Caltech} is a large-scale object recognition dataset that consists of two subsets, ImageNet-1K~\cite{russakovsky2015imagenet} and Caltech256~\cite{griffin2007caltech}.
Here we use images from the public validation set of ImageNet-1K for the target domain. 
In reality, each source domain contains 1,000 and 256 classes, and each target domain contains 84 classes.
	
\textbf{Baseline methods.} We utilize all the source and target samples and report the average classification accuracy and standard deviation over 3 random trials. \textbf{A} $\to$ \textbf{B} means \textbf{A} is the source domain and \textbf{B} is the partial target domain.
For comprehensive comparison, we provide the recognition results of our methods including E-DANN, Ours (w/ BAA) and Ours (BA$^3$US) on each dataset, and compare them with some popular UDA methods \cite{tzeng2017adversarial,long2018conditional} and existing PDA methods, including SAN~\cite{cao2018partialb}, IWAN~\cite{zhang2018importance}, PADA~\cite{cao2018partial}, SSPDA~\cite{bucci2019tackling}, MWPDA~\cite{hu2019multi}, DRCN~\cite{li2020deep}, ETN~\cite{cao2019learning}, and SAFN~\cite{xu2019unsupervised}.

\textbf{Implementation details.} If not specified, all the methods adopt \textbf{ResNet-50}~\cite{he2016deep} as backbone.
We fine-tune the pre-trained ImageNet model in \textbf{PyTorch} using a {NVIDIA Titan X} (12 GB memory). 
The adversarial layer and classifier layer are trained through back-propagation, and the learning rate of the classifier layer is 10 times that of lower layers.
We adopt mini-batch SGD with momentum of 0.9 and the learning rate annealing strategy as \cite{ganin2016domain,long2018conditional}: the learning rate is adjusted by $\eta_p = \eta_0(1 + \hat{\alpha} p)^{-\hat{\beta}}$, where $p$ denotes the training progress changing from 0 to 1, and $\eta_0$ = 0.01, $\hat{\alpha}$ = 10, $\hat{\beta}=$ 0.75 as suggested by \cite{long2018conditional}. 
Based on the number of classes and balancing analysis in \cite{chen2019complement}, we use $\beta=5$ for Office31 and ImageNet-Caltech, and $\beta=1$ for Office-Home.
Besides, we set the number of intervals $N/N_u$ to 10, and $N_u$ is set to 200, 500, and 4,000 for Office31, Office-Home, and ImageNet-Caltech, respectively.
Note that our method does not use the ten-crop technique \cite{cao2018partial,long2018conditional} at the evaluation phase for better performance.

	\setlength{\tabcolsep}{1.0pt}
	\begin{table*}[htbp]
		\centering
		\scriptsize
		\caption{Accuracy (\%) on \textbf{Office-Home} dataset for \emph{partial domain adaptation} via ResNet-50~\cite{he2016deep}. The best in \textbf{\color{red}{bold red}}; the second best in \textit{\color{blue}{italic blue}}.}
		\resizebox{1.0\textwidth}{!}{$
			\begin{tabular}{lcccccccccccca}
			\toprule
			Method & Ar$\to$Cl & Ar$\to$Pr & Ar$\to$Rw & Cl$\to$Ar & Cl$\to$Pr & Cl$\to$Rw & Pr$\to$Ar & Pr$\to$Cl & Pr$\to$Rw & Rw$\to$Ar & Rw$\to$Cl & Rw$\to$Pr & Avg. \\
			\midrule
			ResNet-50~\cite{he2016deep} & 46.33 & 67.51 & 75.87 & 59.14 & 59.94 & 62.73 & 58.22 & 41.79 & 74.88 & 67.40 & 48.18 & 74.17 & 61.35 \\
			ADDA~\cite{tzeng2017adversarial} & 45.23 & 68.79 & 79.21 & 64.56 & 60.01 & 68.29 & 57.56 & 38.89 & 77.45 & 70.28 & 45.23 & 78.32 & 62.82 \\
			CDAN+E~\cite{long2018conditional} & 47.52 & 65.91 & 75.65 & 57.07 & 54.12 & 63.42 & 59.60 & 44.30 & 72.39 & 66.02 & 49.91 & 72.80 & 60.73 \\
			\cmidrule{2-14}
			IWAN~\cite{zhang2018importance} & 53.94 & 54.45 & 78.12 & 61.31 & 47.95 & 63.32 & 54.17 & 52.02 & 81.28 & 76.46 & 56.75 & 82.90 & 63.56 \\
			SAN~\cite{cao2018partialb} & 44.42 & 68.68 & 74.60 & 67.49 & 64.99 & 77.80 & 59.78 & 44.72 & 80.07 & 72.18 & 50.21 & 78.66 & 65.30 \\
			PADA~\cite{cao2018partial} & 51.95 & 67.00 & 78.74 & 52.16 & 53.78 & 59.03 & 52.61 & 43.22 & 78.79 & 73.73 & 56.60 & 77.09 & 62.06 \\
			SSPDA~\cite{bucci2019tackling} & 52.02 & 63.64 & 77.95 & 65.66 & 59.31 & 73.48 & 70.49 & 51.54 & 84.89 & 76.25 &  \textit{\color{blue}60.74} & 80.86 & 68.07 \\
			MWPDA~\cite{hu2019multi} & 55.39 & 77.53 & 81.27 & 57.08 & 61.03 & 62.33 & 68.74 & 56.42 & \textbf{\color{red}86.67} & 76.70 & 57.67 & 80.06 
			& 68.41 \\
			ETN~\cite{cao2019learning} & \textit{\color{blue}59.24} & 77.03 & 79.54 & 62.92 & 65.73 & 75.01 & 68.29 & 55.37 & 84.37 & 75.72 & 57.66 & \textit{\color{blue}84.54} & 70.45 \\
			DRCN~\cite{li2020deep} & 54.00 & 76.40 & 83.00 & 62.10 & 64.50 & 71.00 & 70.80 & 49.80 & 80.50 & 77.50 & 59.10 & 79.90 & 69.00 \\
			\multirow{2}{*}{SAFN~\cite{xu2019unsupervised}} & 58.93 & 76.25 & 81.42 & 70.43 & \textbf{\color{red}72.97} & 77.78 & 72.36 & 55.34 & 80.40 & 75.81 & 60.42 & 79.92 & 71.83 \\
			& $\pm$0.50& $\pm$0.33& $\pm$0.27& $\pm$0.46& $\pm$1.39& $\pm$0.52& $\pm$0.31& $\pm$0.46& $\pm$0.78& $\pm$0.37& $\pm$0.83& $\pm$0.20 &\\
			\midrule\midrule
			\multirow{2}{*}{E-DANN} & 54.05 & 74.12 & 84.06 & 67.06 & 64.95 & 75.15 & 71.29 & 53.09 & 83.42 & 76.00 & 58.17 & 81.53 & {70.24} \\
			& $\pm$0.45 & $\pm$0.12 & $\pm$0.19 & $\pm$0.19 & $\pm$1.10 & $\pm$0.59 & $\pm$0.19 & $\pm$0.37 & $\pm$0.22 & $\pm$0.65 & $\pm$0.51 & $\pm$0.26 & \\
			\cmidrule{2-14}
			\multirow{2}{*}{Ours (w/ BAA)} & 56.20 &  \textit{\color{blue}79.55} &  \textit{\color{blue}86.21} &  \textit{\color{blue}70.86} & 69.94 &  \textit{\color{blue}81.06} &  \textit{\color{blue}72.51} &  \textit{\color{blue}57.91} & 86.47 &  \textit{\color{blue}77.10} & 59.34 & 83.64 & \textit{\color{blue}73.40} \\
			& $\pm$0.28 & $\pm$0.45 & $\pm$0.32 & $\pm$0.67 & $\pm$2.52 & $\pm$0.30 & $\pm$0.32 & $\pm$0.30 & $\pm$0.32 & $\pm$0.62 & $\pm$0.46 & $\pm$0.39 & \\
			\cmidrule{2-14}
			\multirow{2}{*}{Ours (BA$^3$US)} & \textbf{\color{red}60.62} &  \textbf{\color{red}83.16} &  \textbf{\color{red}88.39} &  \textbf{\color{red}71.75} &  \textit{\color{blue}72.79} &  \textbf{\color{red}83.40} &  \textbf{\color{red}75.45} &  \textbf{\color{red}61.59} &  \textit{\color{blue}86.53} &  \textbf{\color{red}79.25} &  \textbf{\color{red}62.80} &  \textbf{\color{red}86.05} & \textbf{\color{red}75.98} \\
			& $\pm$ 0.45 & $\pm$0.12 & $\pm$0.19 & $\pm$0.19 & $\pm$1.10  & $\pm$0.59 & $\pm$0.19 & $\pm$0.37 & $\pm$0.22 & $\pm$0.65 & $\pm$0.51 & $\pm$0.26 & \\
			\bottomrule
			\end{tabular}%
			$}
		\label{tab:oc}%
	\end{table*}%
	
\subsection{Quantitative Results for Partial Domain Adaptation}
The results on three object recognition datasets including \textbf{Office-Home}, \textbf{Office31} and \textbf{ImageNet-Caltech} for PDA are shown in Table~\ref{tab:oc} and \ref{tab:office}, with some baseline results directly reported from ETN~\cite{cao2019learning} with the same protocol.
Obviously, BA$^3$US achieves the best or second-best results on 10 out of 12 transfer tasks on the \textbf{Office-Home} dataset, and Ours (w/ BAA) and SAFN obtain the second and third best results, respectively.
Regarding the average accuracy, BA$^3$US advances the state-of-the-art result on \textbf{Office-Home} in SAFN~\cite{xu2019unsupervised} by 5.78\%, from 71.83\% to 75.98\%.
For two specific tasks \emph{Cl}$\to$\emph{Ar} and \emph{Pr}$\to$\emph{Rw}, BA$^3$US merely performs slightly worse than the best method.
Besides, a state-of-the-art UDA approach CDAN+E~\cite{long2018conditional} performs worse than ResNet-50, which implies the difficulty of the partial transfer task.
Further compared with UDA methods, even PDA methods like IWAN~\cite{zhang2018importance} and PADA~\cite{cao2018partial} do not work well, which again indicates the partial transfer is quite challenging.

On the small-scale \textbf{Office31} dataset, BA$^3$US again obtains the best average accuracy and performs the best in 3 out of 6 transfer tasks. 
For transfer tasks from a large source domain \emph{A} to small target domains (\emph{D}, \emph{W}), BA$^3$US remarkably outperforms other PDA methods.
BA$^3$US performs slightly worse for the \emph{D} $\to$ \emph{A} task because the target domain \emph{D} is very small, making the proposed target augmentation and adaptive uncertainty suppression techniques inefficient. 
On the large-scale \textbf{ImageNet-Caltech} dataset, BA$^3$US performs the best for both transfer tasks and still holds the best average accuracy with significant improvements.
Moreover, UDA methods do not always perform better than ResNet-50, which implies they may suffer from the negative transfer problem.

Besides the ResNet-50 backbone network, we further investigate the effectiveness of BA$^3$US with another backbone network VGG-16~\cite{simonyan2014very} on \textbf{Office31} and compare it with state-of-the-art methods in Table~\ref{tab:vgg}.
It can be clearly observed that BA$^3$US achieves the best or second-best results for all the tasks, significantly advancing the average accuracy from 93.88\% to 95.84\%.
Compared with the results in Table~\ref{tab:office}, we find BA$^3$US (97.81\%$\to$95.84\%) is also robust than ETN (96.73\%$\to$93.88\%) w.r.t. the change of the backbone network.

	\setlength{\tabcolsep}{1.0pt}
	\begin{table*}[htbp]
		\centering
		\scriptsize
		\caption{Accuracy (\%) on \textbf{Office31} and \textbf{ImageNet-Caltech} for \emph{partial domain adaptation} via ResNet-50~\cite{he2016deep}. The best in \textbf{\color{red}{bold red}}; the second best in \textit{\color{blue}{italic blue}}.}
		\resizebox{1.0\textwidth}{!}{$
			\begin{tabular}{lllllllaclla}
			\toprule
			\multirow{2}{*}{Method} &\multicolumn{7}{c}{Office31} & & \multicolumn{3}{c}{ImageNet-Caltech}\\
			\cmidrule{2-8}        \cmidrule{10-12}
			& A $\to$ D & A $\to$ W & D $\to$ A & D $\to$ W & W $\to$ A & W $\to$ D & {Avg.} & & I $\to$ C & C $\to$ I & {Avg.}\\
			\midrule
			ResNet-50~\cite{he2016deep} & 83.44$_{\pm1.12}$ & 75.59$_{\pm1.09}$ & 83.92$_{\pm0.95}$ & 96.27$_{\pm0.85}$ & 84.97$_{\pm0.86}$ & 98.09$_{\pm0.74}$ & 87.05 & & 69.69$_{\pm0.78}$ & 71.29$_{\pm0.74}$ & 70.49 \\
			ADDA~\cite{tzeng2017adversarial} & 83.41$_{\pm0.17}$ & 75.67$_{\pm0.17}$ & 83.62$_{\pm0.14}$ & 95.38$_{\pm0.23}$ & 84.25$_{\pm0.13}$ & \textit{\color{blue}99.85}$_{\pm0.12}$ & 87.03 & & 71.82$_{\pm0.45}$ & 69.32$_{\pm0.41}$ & 70.57\\
			CDAN+E~\cite{long2018conditional} & 77.07$_{\pm0.90}$ & 80.51$_{\pm1.20}$ & 93.58$_{\pm0.07}$ & 98.98$_{\pm0.00}$ & 91.65$_{\pm0.00}$ & 98.09$_{\pm0.00}$ & 89.98 & & 72.45$_{\pm0.07}$ & 72.02$_{\pm0.13}$ & 72.24\\
			\cmidrule{2-12}
			IWAN~\cite{zhang2018importance} & 90.45$_{\pm0.36}$ & 89.15$_{\pm0.37}$ & \textit{\color{blue}95.62}$_{\pm0.29}$ & \textit{\color{blue}99.32}$_{\pm0.32}$ & 94.26$_{\pm0.25}$ & 99.36$_{\pm0.24}$ & 94.69 & & 78.06$_{\pm0.40}$ & 73.33$_{\pm0.46}$ & 75.70\\
			SAN~\cite{cao2018partialb} & 94.27$_{\pm0.28}$ & 93.90$_{\pm0.45}$ & 94.15$_{\pm0.36}$ & \textit{\color{blue}99.32}$_{\pm0.52}$ & 88.73$_{\pm0.44}$ & 99.36$_{\pm0.12}$ & 94.96 & & 77.75$_{\pm0.36}$ & 75.26$_{\pm0.42}$ & 76.51\\
			PADA~\cite{cao2018partial} & 82.17$_{\pm0.37}$ & 86.54$_{\pm0.31}$ & 92.69$_{\pm0.29}$ & \textit{\color{blue}99.32}$_{\pm0.45}$ & \textit{\color{blue}95.41}$_{\pm0.33}$ & \textbf{\color{red}100.0}$_{\pm0.00}$ & 92.69 & & 75.03$_{\pm0.36}$ & 70.48$_{\pm0.44}$ & 72.76\\
			SSPDA~\cite{bucci2019tackling} & 90.87 & 91.52 & 90.61 & 92.88 & 94.36 & 98.94 & 93.20 & & - & - & -\\
			MWPDA~\cite{hu2019multi} & \textit{\color{blue}95.12} & \textit{\color{blue}96.61} &95.02 & \textbf{\color{red}100.0} & \textbf{\color{red}95.51} & \textbf{\color{red}100.0} & 97.05 & & - & - & - \\
			DRCN~\cite{li2020deep} & 86.00 & 88.05 & 95.60 & \textbf{\color{red}100.0} & 95.80 & \textbf{\color{red}100.0} & 94.30 && 75.30 & 78.90 & 77.10 \\
			ETN~\cite{cao2019learning} & 95.03$_{\pm0.22}$ & 94.52$_{\pm0.20}$ & \textbf{\color{red}96.21}$_{\pm0.27}$ & \textbf{\color{red}100.0}$_{\pm0.00}$ & 94.64$_{\pm0.24}$ & \textbf{\color{red}100.0}$_{\pm0.00}$ & 96.73 & & \textit{\color{blue}83.23}$_{\pm0.24}$ & 74.93$_{\pm0.44}$ & 79.08\\
			\midrule\midrule
			E-DANN & 92.36$_{\pm0.00}$ & 93.22$_{\pm0.00}$ & 94.61$_{\pm0.05}$ & \textbf{\color{red}100.0}$_{\pm0.00}$ & 94.71$_{\pm0.05}$ & 98.73$_{\pm0.00}$ & 95.60 & & 78.31$_{\pm0.81}$ & 77.69$_{\pm0.25}$ & 78.00 \\
			Ours (w/ BAA) & 93.63$_{\pm0.00}$ & 93.90$_{\pm0.00}$ & 94.89$_{\pm0.09}$ & \textbf{\color{red}100.0}$_{\pm0.00}$ & 94.78$_{\pm0.00}$ & \textbf{\color{red}100.0}$_{\pm0.00}$ & \textit{\color{blue}96.20} & & 82.97$_{\pm0.49}$ & \textit{\color{blue}79.34}$_{\pm0.08}$ & \textit{\color{blue}81.16} \\
			Ours (BA$^3$US) & \textbf{\color{red}99.36}$_{\pm0.00}$ & \textbf{\color{red}98.98}$_{\pm0.28}$ & 94.82$_{\pm0.05}$ & \textbf{\color{red}100.0}$_{\pm0.00}$ & 94.99$_{\pm0.08}$ & 98.73$_{\pm0.00}$ & \textbf{\color{red}97.81} & & \textbf{\color{red}84.00}$_{\pm0.15}$ & \textbf{\color{red}83.35}$_{\pm0.28}$ & \textbf{\color{red}83.68} \\
			\bottomrule
			\end{tabular}%
			$}
		\label{tab:office}%
	\end{table*}%
	
	\setlength{\tabcolsep}{3.0pt}
	\begin{table*}[htbp]
		\centering
		\scriptsize
		\caption{Accuracy (\%) on \textbf{Office31} for \emph{partial domain adaptation} via VGG-16~\cite{simonyan2014very}. The best in \textbf{\color{red}{bold red}}; the second best in \textit{\color{blue}{italic blue}}.}
		\resizebox{0.92\textwidth}{!}{$
			\begin{tabular}{lcccccca}
			\toprule
			Method & A $\to$ D & A $\to$ W & D $\to$ A & D $\to$ W & W $\to$ A & W $\to$ D & {Avg.} \\
			\midrule
			VGG-16~\cite{simonyan2014very} & 76.43$\pm$0.48 & 60.34$\pm$0.84 & 72.96$\pm$0.56 &  97.97$\pm$0.63 & 79.12$\pm$0.54 & \textit{\color{blue}99.36}$\pm$0.36 & 81.03 \\
			IWAN~\cite{zhang2018importance} & \textit{\color{blue}90.95}$\pm$0.33 & 82.90$\pm$0.31 &  89.57$\pm$0.24 &  79.75$\pm$0.26 & 93.36$\pm$0.22 &  88.53$\pm$0.16 & 87.51 \\
			SAN~\cite{cao2018partialb} & 90.70$\pm$0.20 & 83.39$\pm$0.36 & 87.16$\pm$0.23 &  \textit{\color{blue}99.32}$\pm$0.45 & 91.85$\pm$0.35 & \textbf{\color{red}100.0}$\pm$0.00 & 92.07 \\
			PADA~\cite{cao2018partial} &  81.73$\pm$0.34 & \textit{\color{blue}86.05}$\pm$0.36 & 93.00$\pm$0.24 & \textbf{\color{red}100.0}$\pm$0.00 & \textit{\color{blue}95.26}$\pm$0.27 & \textbf{\color{red}100.0}$\pm$0.00 & 92.54 \\
			ETN~\cite{cao2019learning} &  89.43$\pm$0.17 & 85.66$\pm$0.16 & \textbf{\color{red}95.93}$\pm$0.23 & \textbf{\color{red}100.0}$\pm$0.00 & 92.28$\pm$0.20 & \textbf{\color{red}100.0}$\pm$0.00 & \textit{\color{blue}93.88} \\
			\midrule
			Ours (BA$^3$US) & \textbf{\color{red}95.54}$\pm$0.00 & \textbf{\color{red}89.83}$\pm$0.00 & \textit{\color{blue}94.92}$\pm$0.05 & \textit{\color{blue}99.32}$\pm$0.00 & \textbf{\color{red}95.41}$\pm$0.00 & \textbf{\color{red}100.0}$\pm$0.00 & \textbf{\color{red}95.84}  \\
			\bottomrule
			\end{tabular}%
			$}
		\label{tab:vgg}
	\end{table*}%
	
\textbf{Ablation study.}
As shown in Tables~\ref{tab:oc} and \ref{tab:office}, we provide the results of E-DANN and Ours (w/ BAA) along with BA$^3$US on three datasets.
Ours (w/ BAA) extends E-DANN by using the proposed balanced adversarial alignment technique in Sec.~\ref{sec:baa} instead, while Ours (BA$^3$US) extends Ours (w/ BAA) by considering the proposed adaptive complement entropy objective in Eq.~(\ref{eq:src_coe}).
Firstly, the baseline method E-DANN always performs well for partial domain adaptation since it removes the irrelevant classes in the source classification term like \cite{cao2018partial}.
Secondly, results on all three datasets demonstrate that Ours (BA$^3$US) performs better than Ours (w/ BAA) and Ours (w/ BAA) performs better than E-DANN, which verify the effectiveness of two proposed techniques in Sec.~\ref{sec:baa} and Sec.~\ref{sec:aus}.
	
\subsection{Quantitative Results for Closed-set Domain Adaptation} 
This section further investigates the effectiveness of the proposed uncertainty suppression technique for vanilla closed-set domain adaptation.
Here we consider integrating them with DANN and CDAN~\cite{long2018conditional} respectively, and compare our methods with state-of-the-art UDA approaches including \cite{wang2019transferable,xu2019unsupervised} on the most-favored \textbf{Office-Home} dataset.
As shown in Table~\ref{tab:uda}, the proposed method BA$^3$US built on CDAN obtains the best average accuracy and ranks the top two in 9 out of 12 different transfer tasks.
It is obvious that the adaptive uncertainty suppression technique works well for closed-set domain adaptation, advancing the average accuracy from 67.6\% to 68.7\% and from 68.0\% to 69.2\%.
In fact, we also study the effectiveness of balanced adversarial alignment but find it hardly improve the performance since the label distributions in UDA have already been symmetric.
	
	\setlength{\tabcolsep}{1.0pt}
	\begin{table*}[htbp]
		\centering
		\small
		\caption{Accuracy (\%) on \textbf{Office-Home} dataset for \emph{vanilla unsupervised domain adaptation} via ResNet-50~\cite{he2016deep}. Methods$^*$ utilize augmentation during evaluation.}
		\resizebox{0.92\textwidth}{!}{$
			\begin{tabular}{lcccccccccccca}
			\toprule
			Method & Ar$\to$Cl & Ar$\to$Pr & Ar$\to$Rw & Cl$\to$Ar & Cl$\to$Pr & Cl$\to$Rw & Pr$\to$Ar & Pr$\to$Cl & Pr$\to$Rw & Rw$\to$Ar & Rw$\to$Cl & Rw$\to$Pr & Avg. \\
			\midrule
			CDAN+E~\cite{long2018conditional}$^*$ & 50.7  & 70.6  & 76.0  & 57.6  & 70.0  & 70.0  & 57.4  & 50.9  & 77.3  & 70.9  & 56.7  & 81.6  & 65.8 \\
			MCS~\cite{liang2019distant} & 55.9 & \textit{\color{blue}73.8} & \textbf{\color{red}79.0} & 57.5 & 69.9 & 71.3 & 58.4 & 50.3 & 78.2 & 65.9 & 53.2 & 82.2 & 66.3 \\
			DRCN~\cite{li2020deep} & 50.6 & 72.4 & 76.8 & 61.9 & 69.5 & 71.3 & 60.4 & 48.6 & 76.8 & 72.9 & 56.1 & 81.4 & 66.6 \\
			CDAN+TransNorm~\cite{wang2019transferable}$^*$  & 50.2 & 71.4 & 77.4 & 59.3 & 72.7 & 73.1 & 61.0 & 53.1 & 79.5 & 71.9 & \textbf{\color{red}59.0} & 82.9 & 67.6 \\
			SAFN~\cite{xu2019unsupervised}$^*$ & \textbf{\color{red}54.4} & 73.3 & 77.9 & \textbf{\color{red}65.2} & 71.5 & 73.2 & \textit{\color{blue}63.6} & 52.6 & 78.2 & 72.3 & 58.0 & 82.1 & 68.5 \\
			SAFN~\cite{xu2019unsupervised}  & 52.0 & 71.7 & 76.3 & \textit{\color{blue}64.2} & 69.9 & 71.9 & \textbf{\color{red}63.7} & 51.4 & 77.1 & 70.9 & 57.1 & 81.5 & 67.3 \\
			\midrule\midrule
			Ours (w/ BAA) &  50.9 & 72.0 & 77.5 & 61.2 & 72.6 & 72.7 & 62.8 & 52.7 & 79.9 & 70.8 & 56.6 & 82.7 & 67.6 \\
			Ours (BA$^3$US) & 51.2 & \textit{\color{blue}73.8} & \textit{\color{blue}78.1} & 63.3 & \textit{\color{blue}73.4} & \textit{\color{blue}73.6} & 63.3 & \textit{\color{blue}54.5} & \textbf{\color{red}80.4} & \textit{\color{blue}72.6} & 56.7 & \textbf{\color{red}83.7} & \textit{\color{blue}68.7} \\
			\midrule
			Ours (w/ BAA)+CDAN & 52.2 & 73.1 & 77.9 & 61.4 & 72.7 & 73.2 & 61.0  & 51.8 & \textit{\color{blue}80.0} & 72.0 & 57.8 & 83.3 & 68.0 \\
			Ours (BA$^3$US)+CDAN & \textit{\color{blue}54.1} & \textbf{\color{red}74.2} & 77.7 & 62.9 & \textbf{\color{red}73.6} & \textbf{\color{red}74.6} & 63.4 & \textbf{\color{red}54.9} & \textbf{\color{red}80.4} & \textbf{\color{red}73.1} & \textit{\color{blue}58.2} & \textit{\color{blue}83.6} & \textbf{\color{red}69.2} \\
			\bottomrule
			\end{tabular}%
			$}
		\label{tab:uda}%
	\end{table*}%
	
\subsection{Qualitative Results for Partial Domain Adaptation} 
\begin{figure}[h]
		\centering
		\scriptsize
		\renewcommand\arraystretch{1.0}
		\begin{tabular}{ccc}
			\includegraphics[width=0.32\linewidth, trim=65 200 75 235,clip]{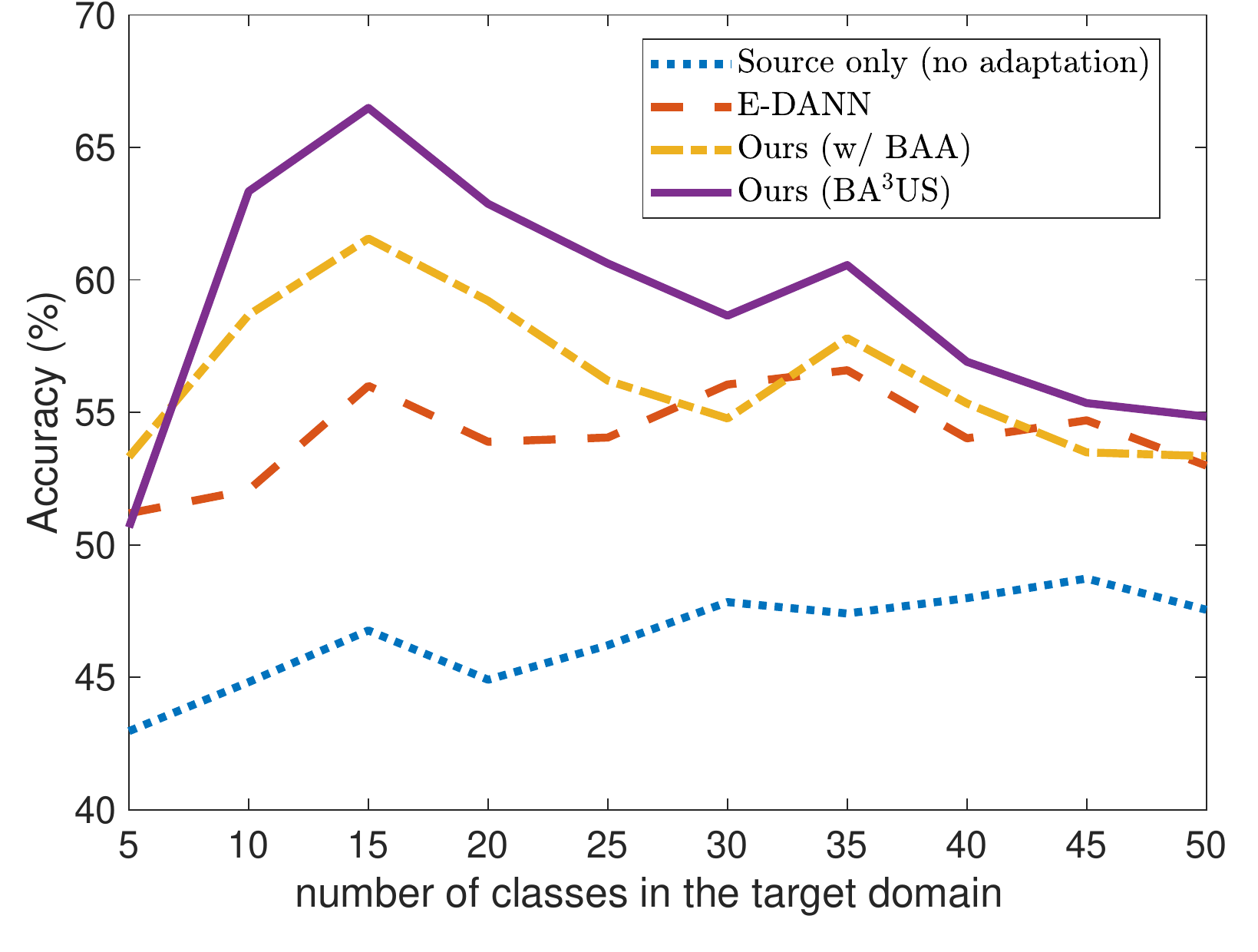} & 
			\includegraphics[width=0.32\linewidth, trim=65 200 75 235,clip]{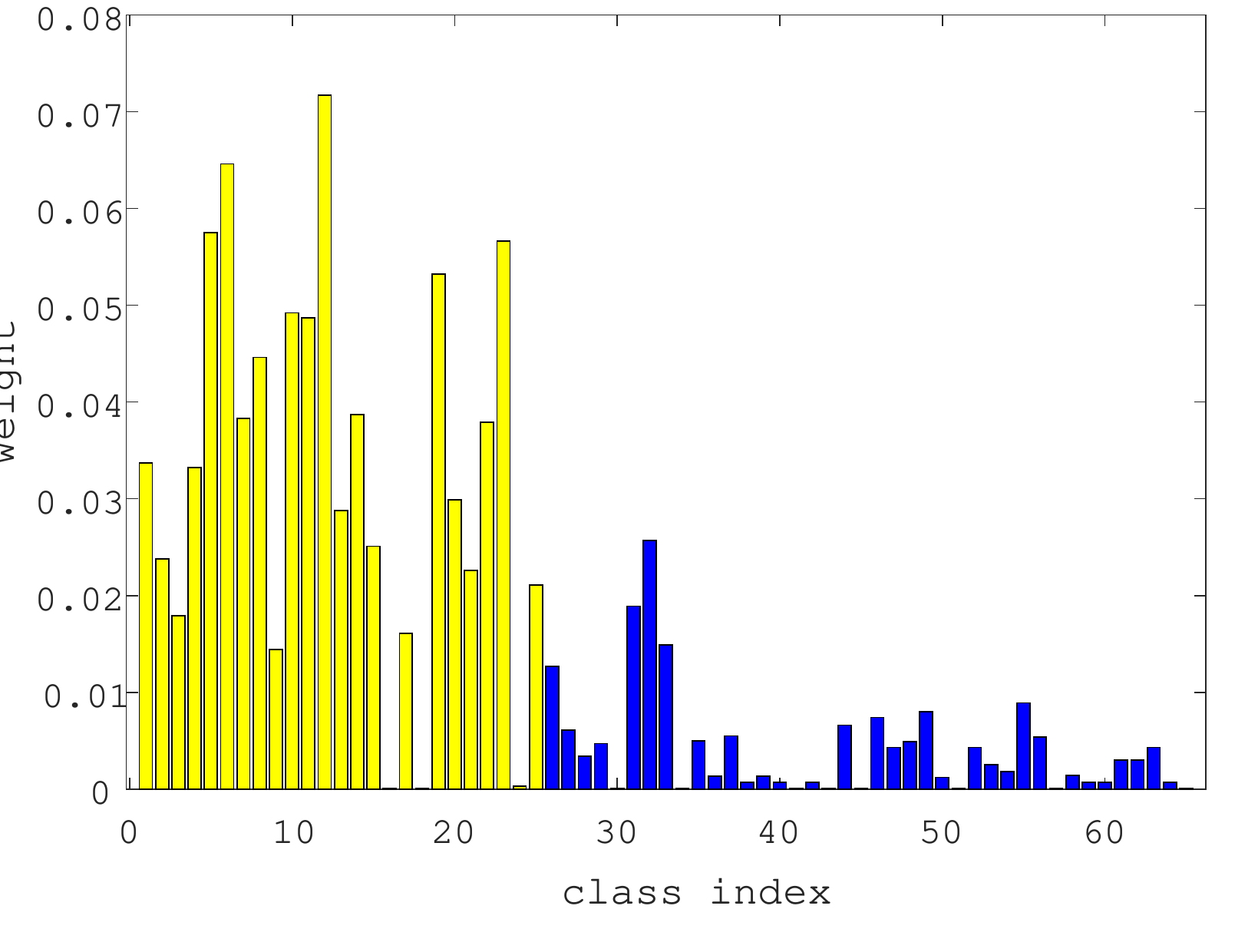} &
			\includegraphics[width=0.32\linewidth, trim=65 200 75 235,clip]{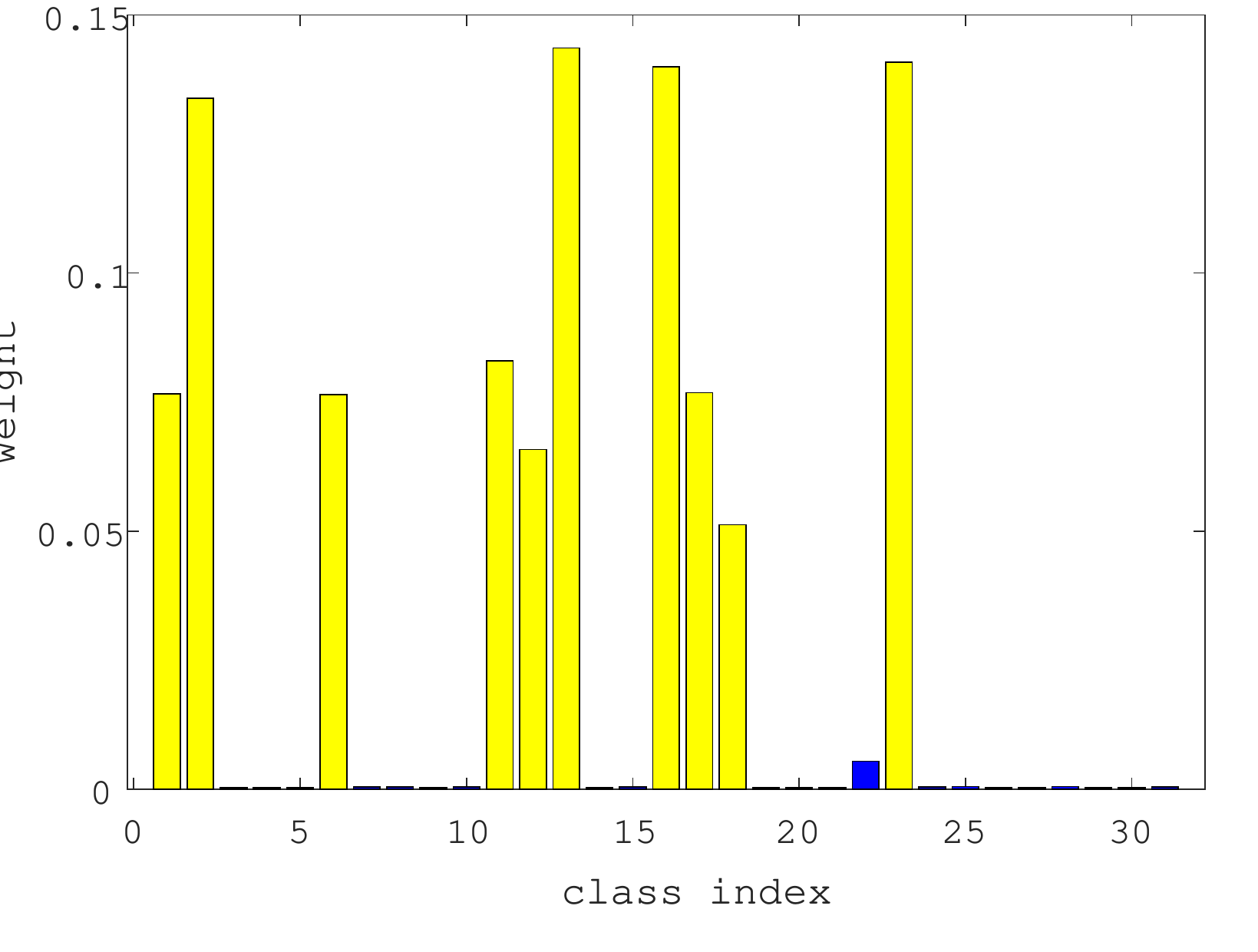}\\
			(a) Accuracy of Ar$\to$Cl & (b) estimated weight of Ar$\to$Cl & (c) estimated weight of A$\to$D
		\end{tabular}
		\caption{(a) Accuracy with varying number of target classes. (b-c) Estimated class-level weights. Yellow bins denote ground-truth classes. Best viewed in color.}
		\label{fig:par}
	\end{figure}

We study our methods with different numbers of target classes in Fig.~\ref{fig:par}(a).
The performance decreases when the number is larger than 15, and BA$^3$US always obtains the best results.
As expected, the proposed augmentation technique becomes important when the number of target classes is small.

\textbf{Weight visualization.}
As stated before, the weighting scheme plays an important role in PDA.  
Thus we investigate the quality of the estimated class-level weight $m$ in Algorithm~\ref{alg}.
As shown in Fig.~\ref{fig:par}(b-c), we plot the estimated weights of BA$^3$US for the two specific tasks.
Since the \emph{Cl} domain and \emph{Ar} domain are quite different, making the estimation challenging. This is also evidenced in other PDA methods in Table~\ref{tab:oc} since the accuracy is around 55\%. 
For the relatively easy task A$\to$D, the weight estimation seems accurate, resulting in high classification accuracy.
	
\textbf{Parameter Sensitivity.} We study the sensitivity of parameters $\xi$ and $\beta$ of BA$^3$US on the \textbf{Office-Home} and \textbf{Office31} datasets.
The mean accuracy is reported in Table~\ref{tab:p1} and Table~\ref{tab:p2}, respectively.
Note that the complement entropy~\cite{chen2019complement} can be considered as a special case of the proposed adaptive one where $\xi=0$.
The results indicate using an adaptive objective is much better, and the performance is relatively stable.
Regarding the parameter $\beta$, the accuracy around 1.0 and 5.0 is also stable, implying our method is not sensitive.
	
	\setlength{\tabcolsep}{3.0pt}
	\begin{table}[h]
		\centering
		\begin{minipage}{0.48\linewidth}
			\centering
			\caption{Sensitivity of parameter $\xi$.}
			\resizebox{0.95\textwidth}{!}{$
				\begin{tabular}{lccccccc}
				\toprule
				Avg. (\%) & 0.0~\cite{chen2019complement}  & 0.1 & 0.3 & 0.5 & 0.7 & 0.9 & 1.0 \\
				\midrule 
				\textbf{Office-Home} & 75.32 & 75.88 & \textbf{\color{red}76.28} & \textit{\color{blue}76.10} & \textit{\color{blue}76.10} & 75.81 & 75.98 \\  
				\textbf{Office31} & 97.68 & 97.71 & 97.65 & 97.67 & 97.64 & \textbf{\color{red}97.84} & \textit{\color{blue}97.81} \\  
				\bottomrule
				\end{tabular}
				$}
			\label{tab:p1}
		\end{minipage}
		\begin{minipage}{0.48\linewidth}
			\centering
			\caption{Sensitivity of parameter $\beta$.}
			\resizebox{0.82\textwidth}{!}{$
				\begin{tabular}{lcccccc}
				\toprule
				Avg. (\%) & 0.0 & 0.1 & 0.5 & 1.0 & 5.0 & 10.0 \\
				\midrule 
				\textbf{Office-Home} & 73.40 & 73.58 & 75.09 & \textbf{\color{red}75.98} & \textit{\color{blue}75.69} & 73.25 \\
				\textbf{Office31} & 96.20 & 96.13 & 96.50 & 96.63 & \textit{\color{blue}97.81} & \textbf{\color{red}97.83} \\   
				\bottomrule
				\end{tabular}
				$}
			\label{tab:p2}
		\end{minipage}
	\end{table}
	
\begin{figure}[h]
		\centering
		\scriptsize
		\setlength\tabcolsep{0mm}
		\renewcommand\arraystretch{1.0}
		\begin{tabular}{cc}
			\includegraphics[width=0.4\linewidth, trim=65 200 75 235,clip]{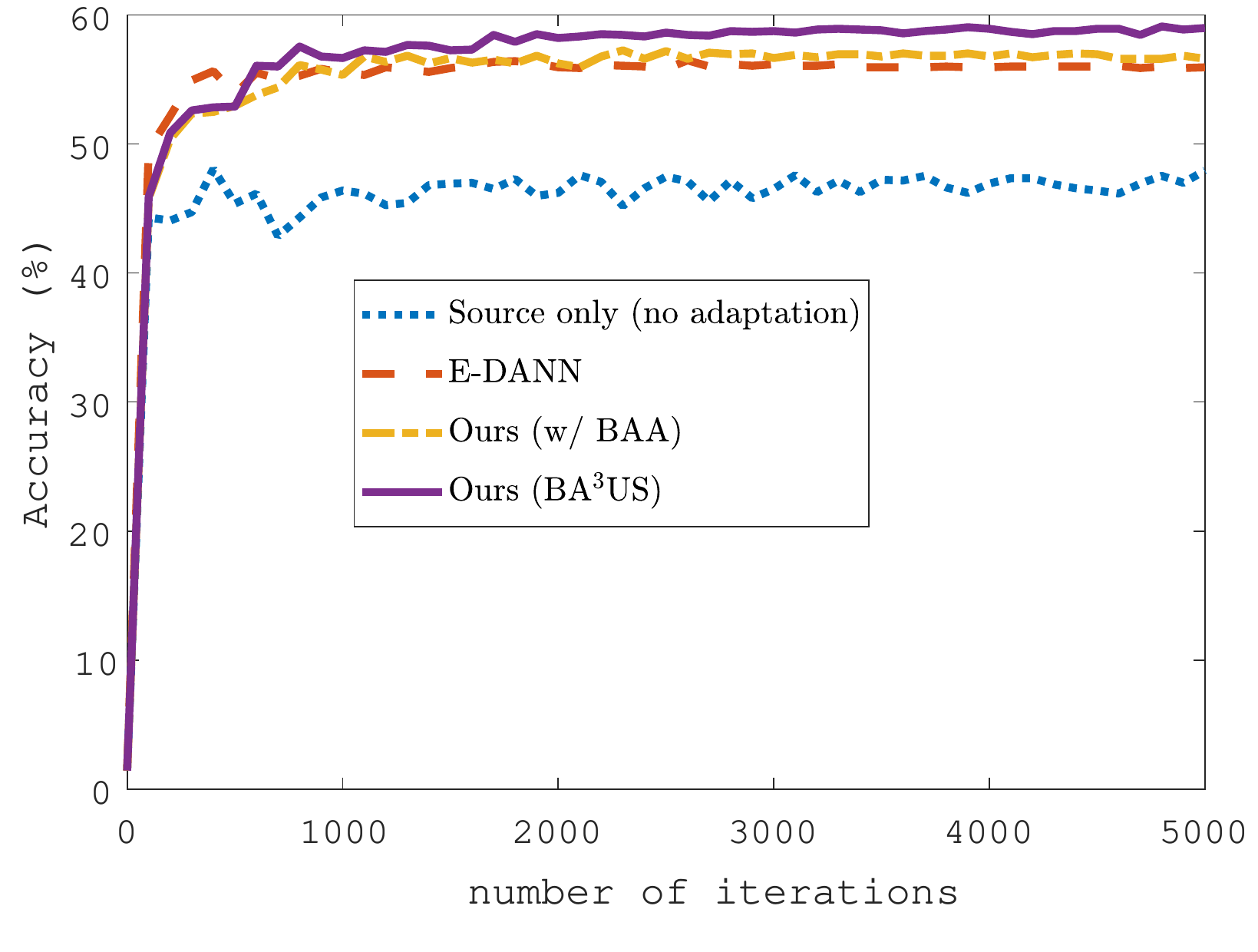} & 
			\includegraphics[width=0.4\linewidth, trim=65 200 75 235,clip]{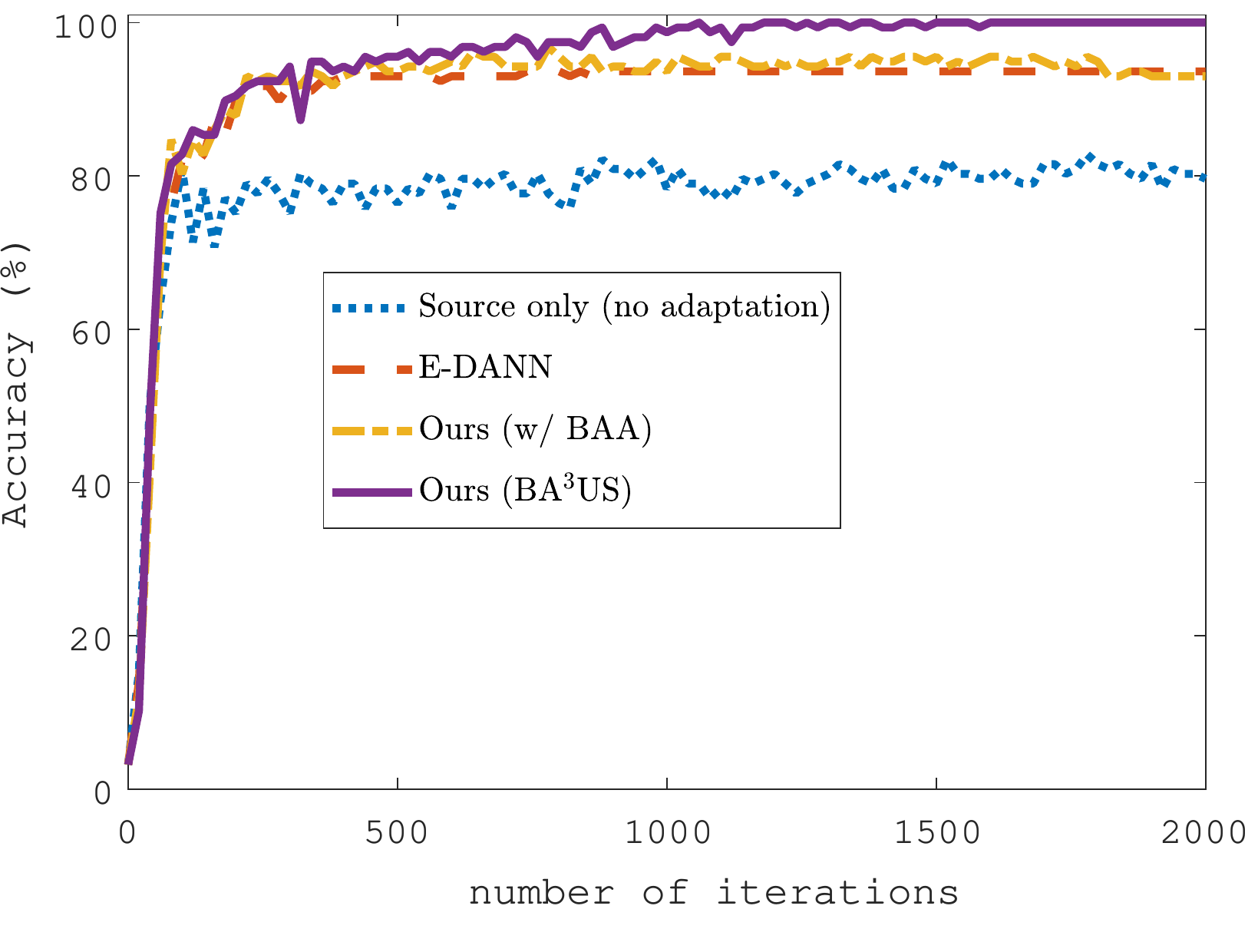}\\
			(a) Ar $\to$ Cl (Office-Home) & (b) A $\to$ D (Office31)
		\end{tabular}
		\caption{Convergence analysis of proposed methods on two different transfer tasks. Accuracy (\%) is given w.r.t. number of iterations. Best viewed in color.}
		\label{fig:convergence}
	\end{figure}

\textbf{Convergence performance.}
As shown in Fig.~\ref{fig:convergence}, we study the convergence performance of the proposed methods for Ar$\to$Cl and A$\to$D.
Obviously, the `source only' method works worse without the domain alignment module, and E-DANN performs much better than it and quickly converges after 1,000 iterations.
Besides, both BA$^3$US and Ours (w/ BAA) obtain similar promising results, and BA$^3$US performs slightly better since it further considers the adaptive complement entropy to diminish the uncertainty in source predictions.
	
	\begin{figure}[!h]
		\centering
		\footnotesize
		\setlength\tabcolsep{0mm}
		\renewcommand\arraystretch{0.1}
		\begin{tabular}{cccc}
			\includegraphics[width=0.24\linewidth, trim=35 30 40 35,clip]{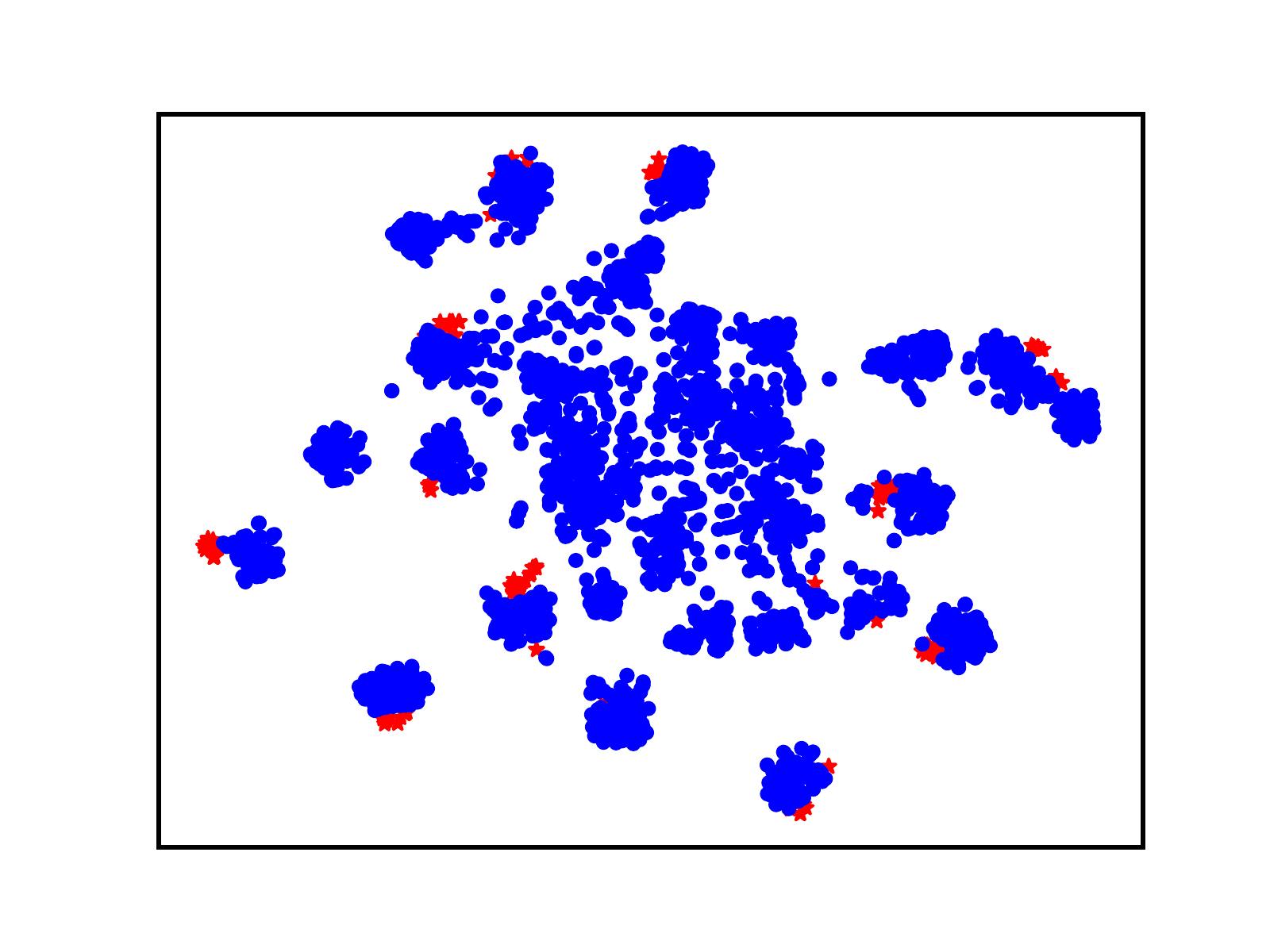} &
			\includegraphics[width=0.24\linewidth, trim=35 30 40 35,clip]{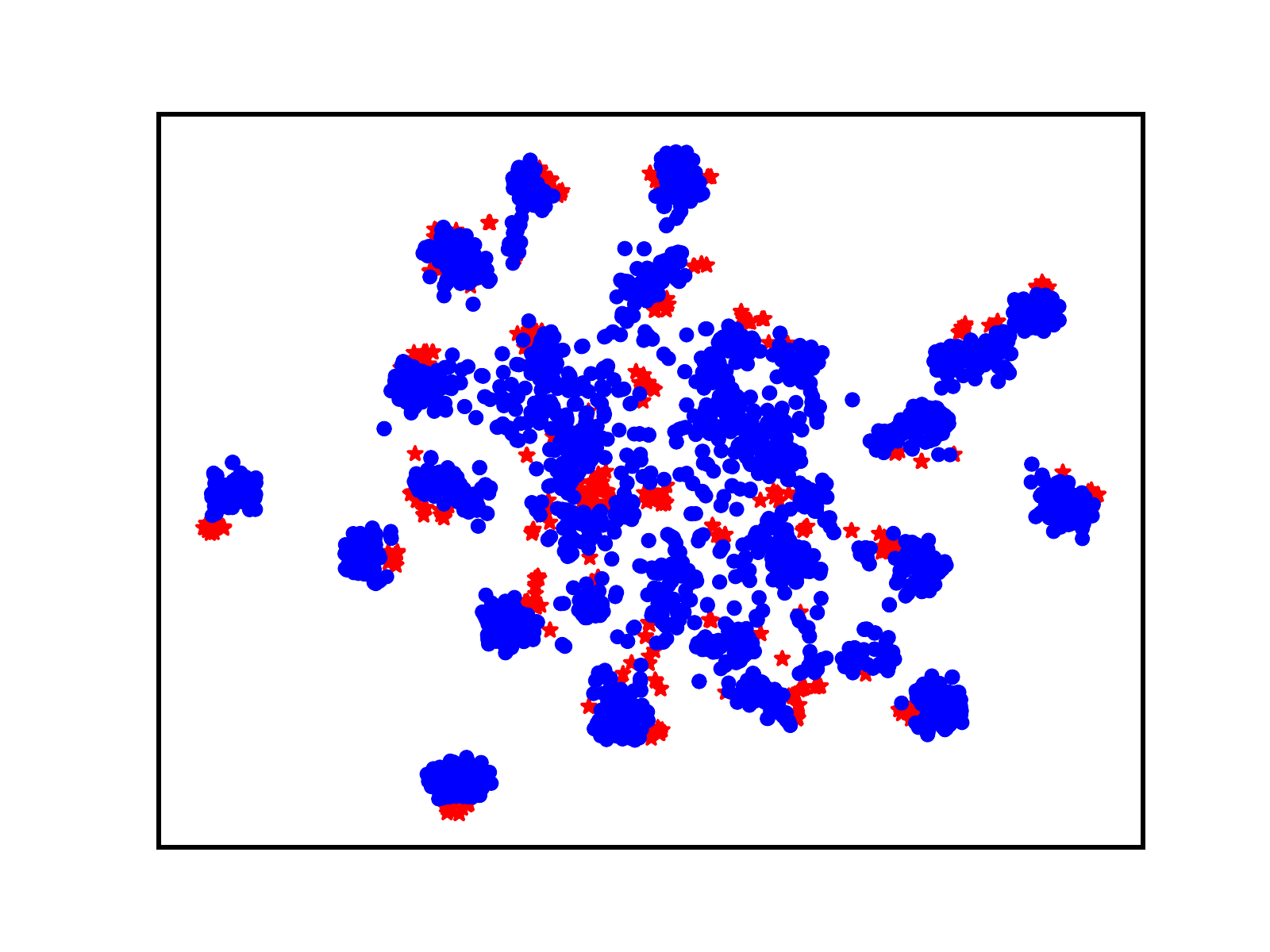} & 
			\includegraphics[width=0.24\linewidth, trim=35 30 40 35,clip]{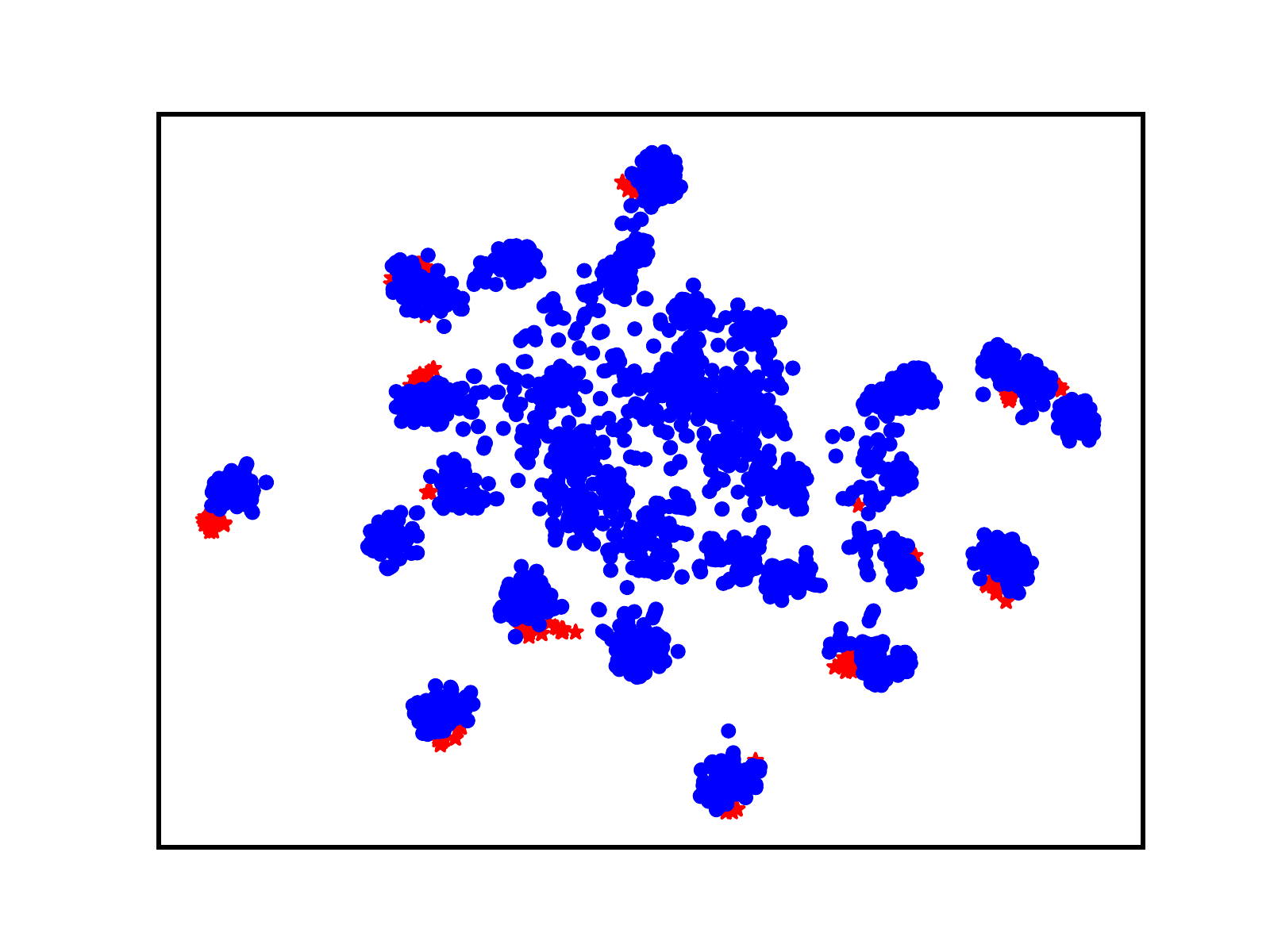} &
			\includegraphics[width=0.24\linewidth, trim=35 30 40 35,clip]{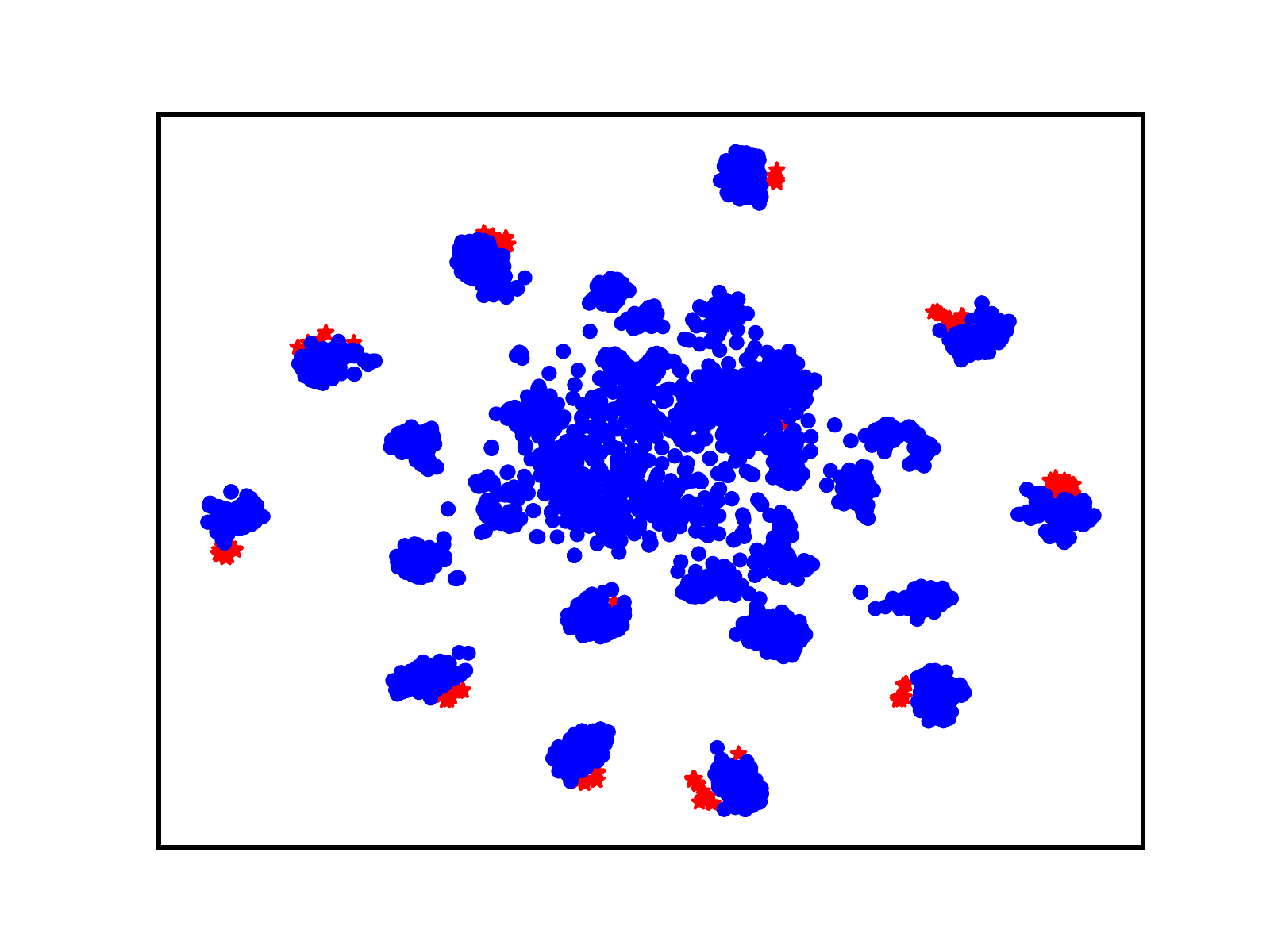}\\
			\includegraphics[width=0.24\linewidth, trim=35 30 40 35,clip]{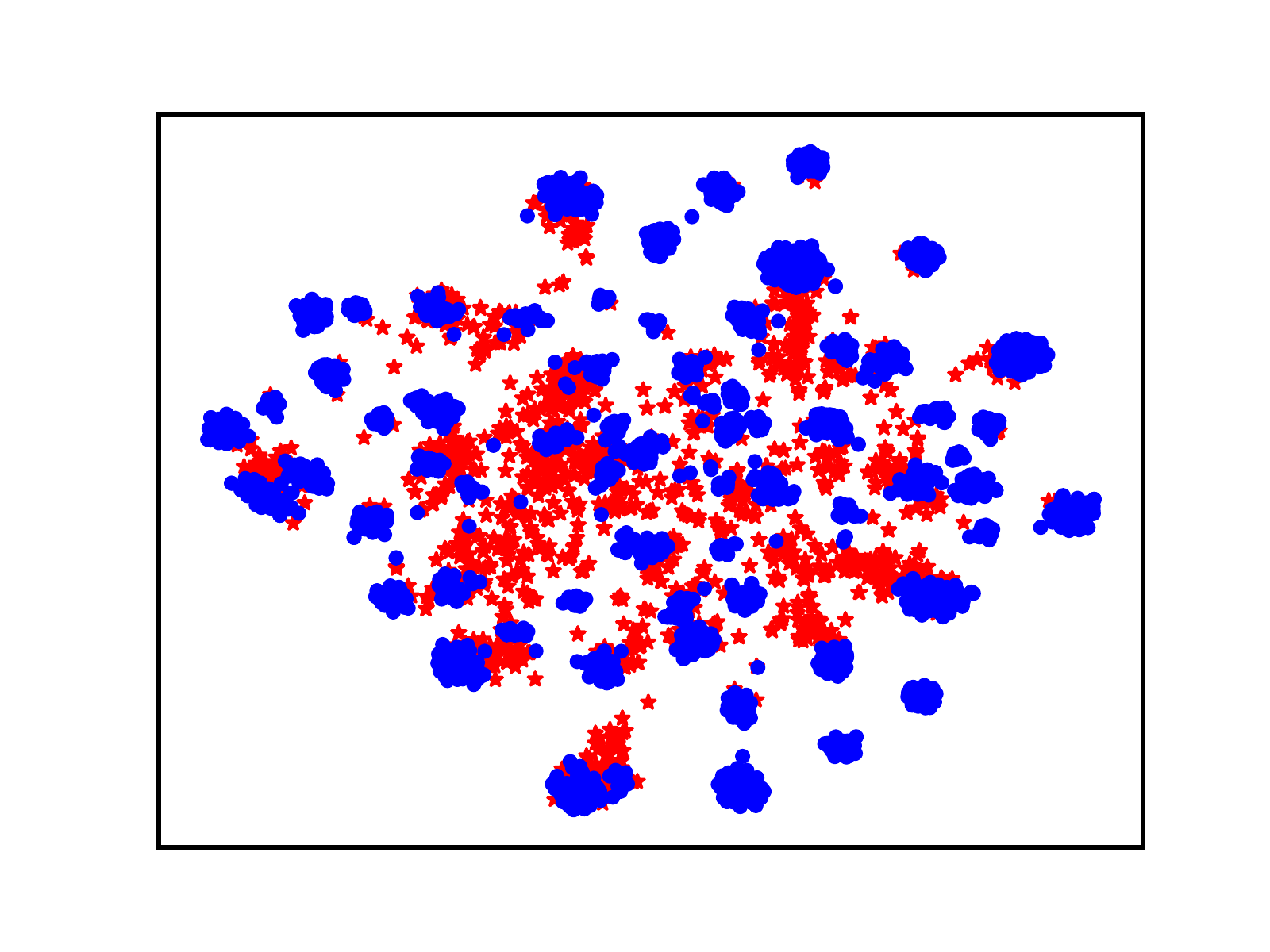} &
			\includegraphics[width=0.24\linewidth, trim=35 30 40 35,clip]{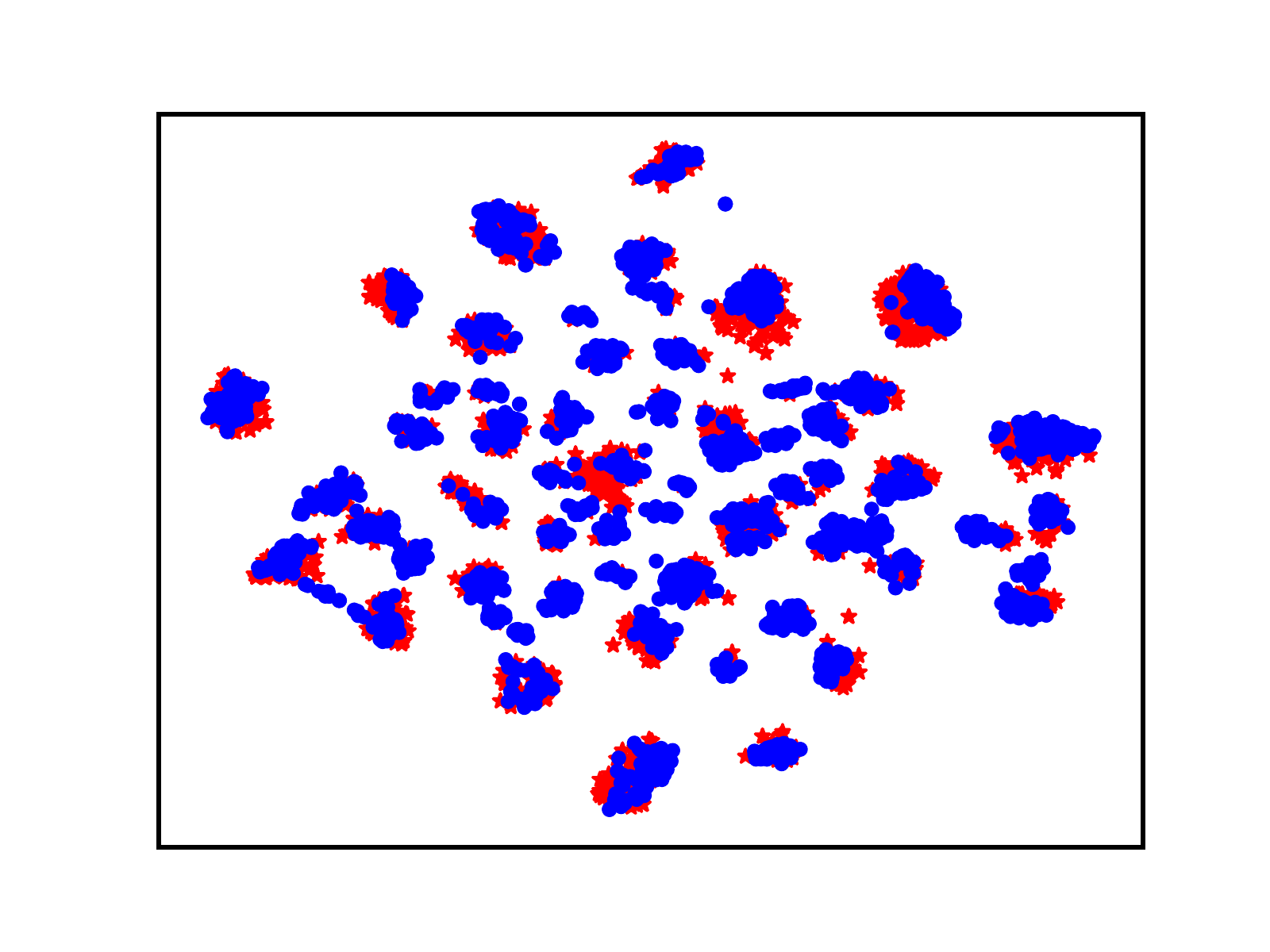} & 
			\includegraphics[width=0.24\linewidth, trim=35 30 40 35,clip]{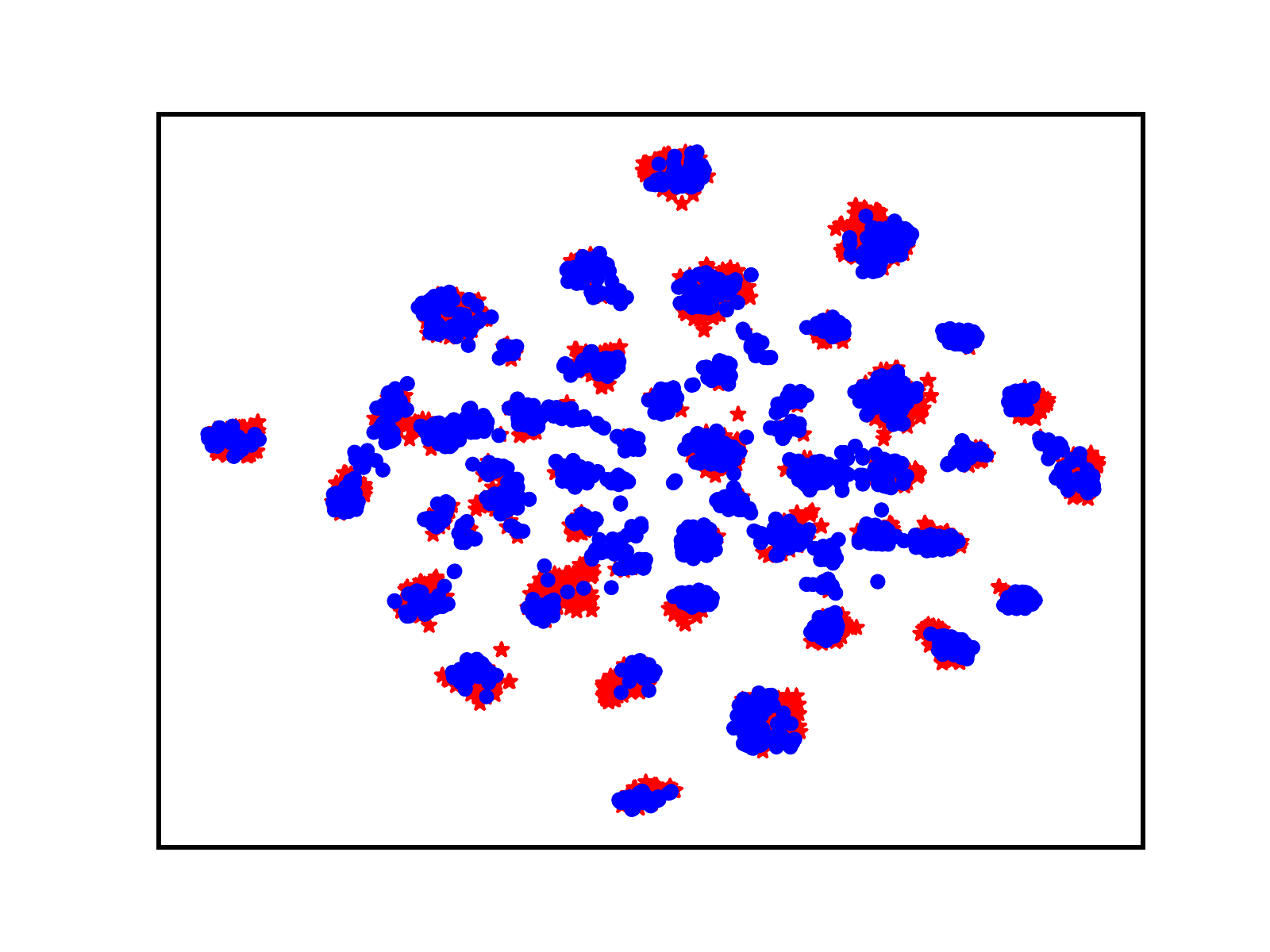} &
			\includegraphics[width=0.24\linewidth, trim=35 30 40 35,clip]{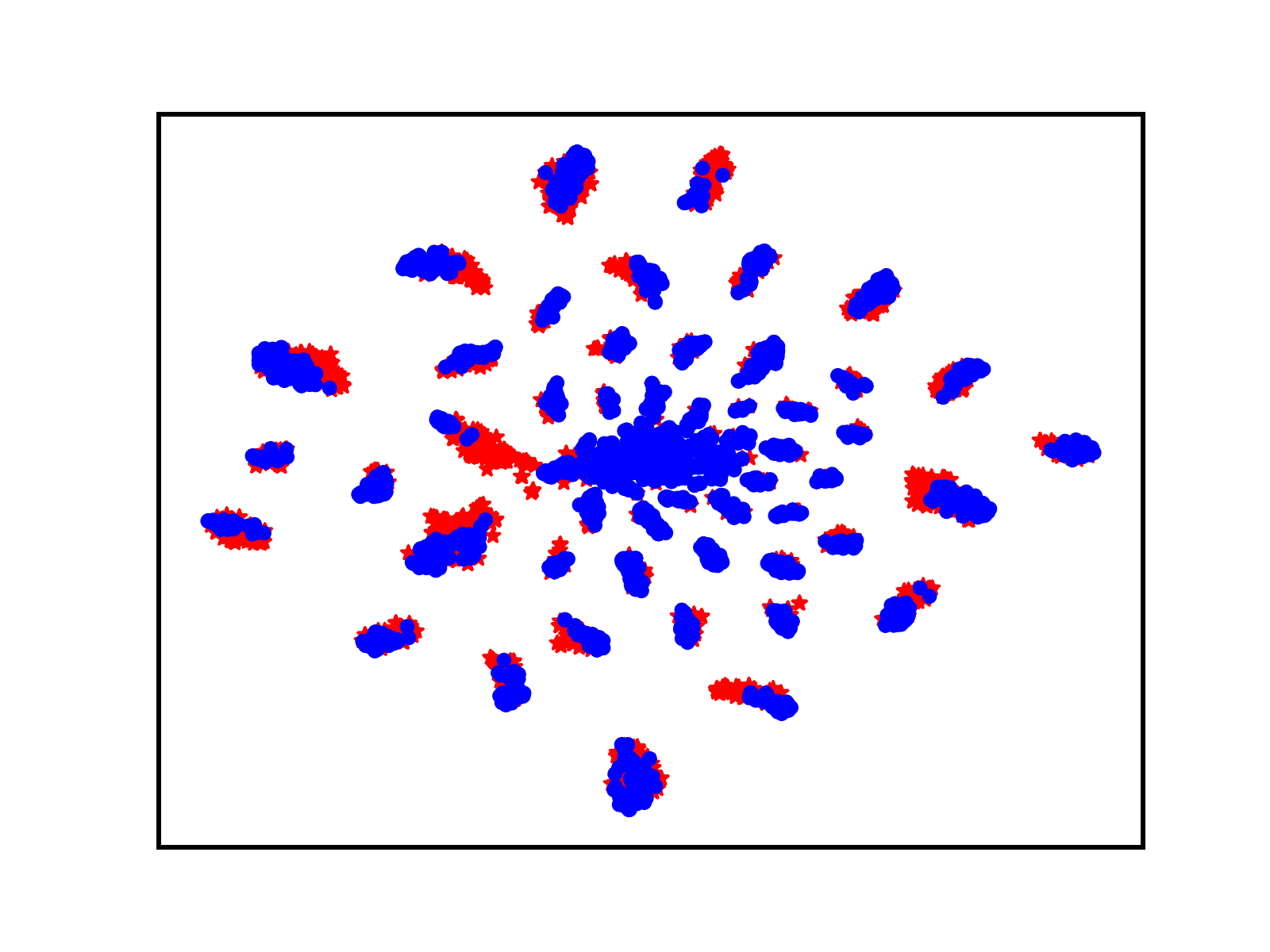}\\
			~\\
			(a) Source only & (b) E-DANN & (c) Ours (w/ BAA) & (d) Ours (BA$^3$US) \\
		\end{tabular}
		\caption{t-SNE visualizations for two transfer tasks A$\to$D (upper row) and Ar$\to$Cl (bottom row). {\color{blue}Blue: source data}; {\color{red}red: target data}.}
		\label{fig:tsne}
	\end{figure}
	
\textbf{Feature visualization.}
We plot in Fig.~\ref{fig:tsne} the t-SNE embeddings \cite{maaten2008visualizing} of the features learned by `source only', E-DANN and BA$^3$US for two different transfer tasks.
It is easy to discover that the features of target data are rather confusing in `no adaptation' while the balanced alignment module helps mitigate the domain gap and the adaptive uncertainty suppression module helps increase the discrimination of the features.
	
\section{Conclusion}
We develop a novel adversarial learning-based method BA$^3$US for partial domain adaptation, which well addresses two key problems, negative transfer and uncertainty propagation.
To tackle the asymmetric label distributions, BA$^3$US  offers a very simple solution by augmenting the target domain with samples from the source domain.
Then, it uncovers an overlooked issue in the field termed uncertainty propagation and designs an adaptive complement entropy objective to well suppress the uncertainty in source predictions.
Empirical results show it also works well for vanilla closed-set domain adaptation. 
Further experiments have validated that BA$^3$US improves existing methods with substantial gains, establishing new state-of-the-art. 

\begin{algorithm}
	\footnotesize
	\renewcommand\baselinestretch{0.4}\selectfont
	\SetAlgoLined
	\textbf{Input}: Labeled source domain $\mathcal{D}_s$, unlabeled target domain $\mathcal{D}_t$\;
	\textbf{Parameters}: Total training iterations $N$, updating interval $N_u$, batch size $B_s=36$, $\rho_0=1/4$, $\xi=1$, $\alpha=0.1$, $\beta \in \{1,5\}$, $\lambda$\; 
	Initialize the model parameters $\theta_f, \theta_g, \theta_d$\;
	Initialize the class-level weight vector $m$, $m_i=1/C$\;
	\For{i = 1 to $N$}{
		Obtain $B_s$ samples from $\mathcal{D}_s$ and $\mathcal{D}_t$, respectively\;
		Obtain $\rho B_s$ random samples from $\mathcal{D}_s$\;
		Update $\theta_f, \theta_g, \theta_d$ by optimizing Eq.~(\ref{ours}) and gradient reversaral layer\;
		\If{i \% $N_u$ == 0}{
			update the class-level weight vector $m$\; \textcolor{gray}{\% note that this step is ignored in closed-set domain adaptation}\\
			calcuate $\mathcal{L}_{ent}$ in Eq.~(\ref{tar_ent}) for model selection\;
			$\rho$ $\leftarrow$ $\rho_0$ (1 - $N_u$/$N$)\;
		}
	}
	\textbf{Output}: Target outputs corresponding to the minimal value of $\mathcal{L}_{ent}$.
	\caption{Pseudo code of our method termed BA$^3$US.}
	\label{alg}
\end{algorithm}

%

\bibliographystyle{splncs04}
\bibliography{egbib}
	
\end{document}